\documentclass[11pt,a4paper]{article}

\usepackage[utf8]{inputenc}
\usepackage{multicol}
\usepackage{multirow}

\usepackage{floatpag}
\usepackage{wrapfig, floatrow}
\usepackage[font=small]{caption}
\usepackage{float}
\usepackage[round,authoryear]{natbib}
\usepackage{afterpage}

\usepackage{bm}
\usepackage{amsmath}

\usepackage[normalem]{ulem} 

\usepackage{boxedminipage}

\usepackage[textwidth=1in]{todonotes}

\usepackage{enumitem}

\usepackage{listings}
\usepackage{color}
\definecolor{codegreen}{rgb}{0,0.5,0}
\definecolor{codeblue}{rgb}{0,0,0.9}
\definecolor{codegray}{rgb}{0.5,0.5,0.5}
\definecolor{codepurple}{rgb}{0.58,0,0.82}
\definecolor{backcolour}{rgb}{0.95,0.95,0.92}
\definecolor{backcolour2}{rgb}{0.9,0.9,0.9}
\definecolor{codered}{rgb}{0.5,0,0}
\definecolor{textcodered}{rgb}{0.4,0,0}
\definecolor{palegray}{rgb}{0.98,0.98,0.99}

\lstdefinestyle{mystyle}{
    backgroundcolor=\color{backcolour},   
    commentstyle=\color{codered},
    keywordstyle=\color{codeblue},
    numberstyle=\tiny\color{codegray},
    stringstyle=\color{codegreen},
    breakatwhitespace=false,         
    breaklines=true,                 
    captionpos=b,                    
    keepspaces=true,                 
    numbersep=5pt,                  
    showspaces=false,                
    showstringspaces=false,
    showtabs=false,                  
    tabsize=2,
    otherkeywords={with},
    basicstyle=\ttfamily\footnotesize
}

\usepackage{textcomp}

\newcommand{\mono}[1]{\texttt{\color{textcodered}#1}}

\usepackage{ifpdf}

\usepackage[T1]{fontenc}
\usepackage{fancyhdr}   
\usepackage{makeidx}

\usepackage{lipsum}
\usepackage{verbatim}
\usepackage{lmodern}

\usepackage{booktabs} 

\usepackage{rotating}
\usepackage{blindtext}
\usepackage[includemp=false, marginparwidth=1.8cm, hmargin=3.5cm, vmargin=3cm]{geometry}
\ifnum\pdfoutput>0			
\usepackage[pdftex, 			
pagebackref = false, 			
bookmarks = true, 			
bookmarksnumbered = false, 	
hyperindex = true,			
linktocpage = true,			
pdfpagemode = UseNone, 		
pdfstartview = Fit, 			
pdfpagelayout = OneColumn,
colorlinks = true, 			
urlcolor = blue, 				
citecolor = blue,
linkcolor = blue,
pdfborder = {0 0 0} 			
]{hyperref} 					
\fi

\ifpdf
\DeclareGraphicsExtensions{.png, .pdf, .jpg, .tif}
\else
\DeclareGraphicsExtensions{.eps, .jpg}
\fi

\usepackage{amsmath, amsthm}
\usepackage{amsfonts}
\usepackage{amssymb}

\defcitealias{cmu_mocap}{CMU Motion Capture Database}

\definecolor{lightgrey}{rgb}{.5,.7,.6}

\newcommand{\myfigure}[2]{
\begin{figure}[H]
\floatbox[{\capbeside\thisfloatsetup{capbesideposition={right,top},
capbesidewidth=0.79\textwidth}}]{figure}[\FBwidth]
{\caption*{#1}}
{\includegraphics[width=2.5cm]{#2}}
\end{figure}
\vspace{-.5cm}
}

\graphicspath{{figures/}}

\title{{\bf \huge DeepMind Control Suite}}

\author{ 
\small Yuval Tassa, Yotam Doron, Alistair Muldal, Tom Erez,\\
\small Yazhe Li, Diego de Las Casas, David Budden, Abbas Abdolmaleki, Josh Merel,\\
\small Andrew Lefrancq, Timothy Lillicrap, Martin Riedmiller
}
\date{\small \today}

\begin{document}

\maketitle
\vspace{-.6cm}

\begin{figure*}[ht]
\centering
\begin{minipage}[c]{1\textwidth}
\def\mywidth{1.9cm}
\def\myhsep{-2mm}
\includegraphics[width=\mywidth]{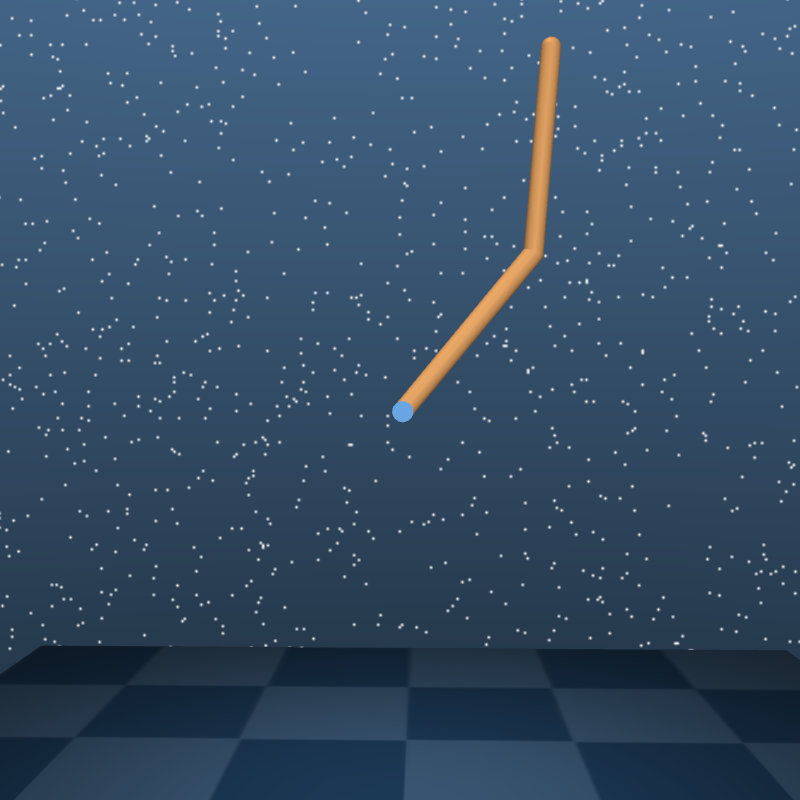}
\hspace{\myhsep}
\includegraphics[width=\mywidth]{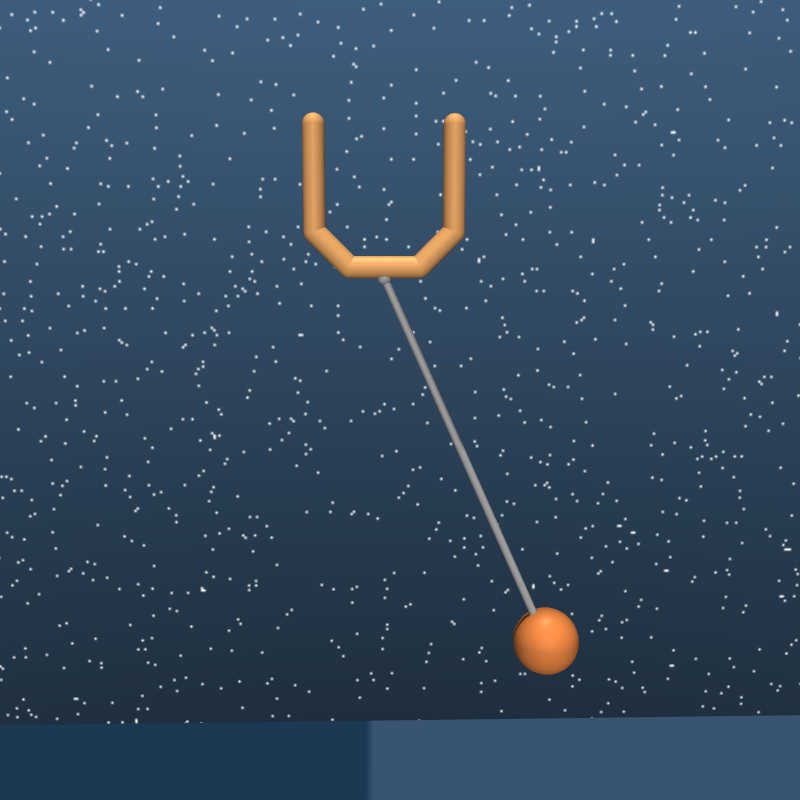}
\hspace{\myhsep}
\includegraphics[width=\mywidth]{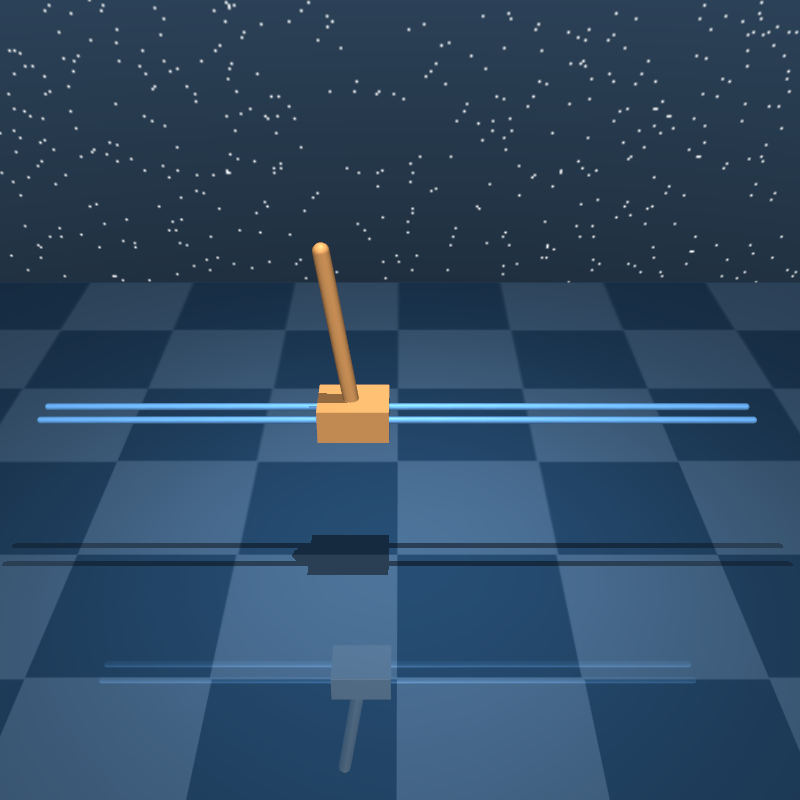}
\hspace{\myhsep}
\includegraphics[width=\mywidth]{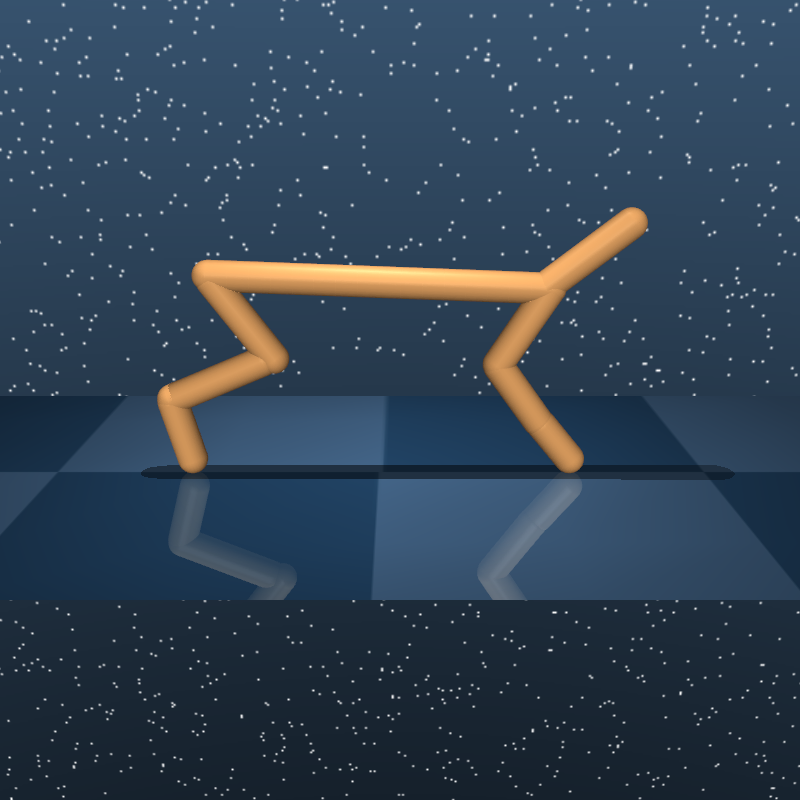}
\hspace{\myhsep}
\includegraphics[width=\mywidth]{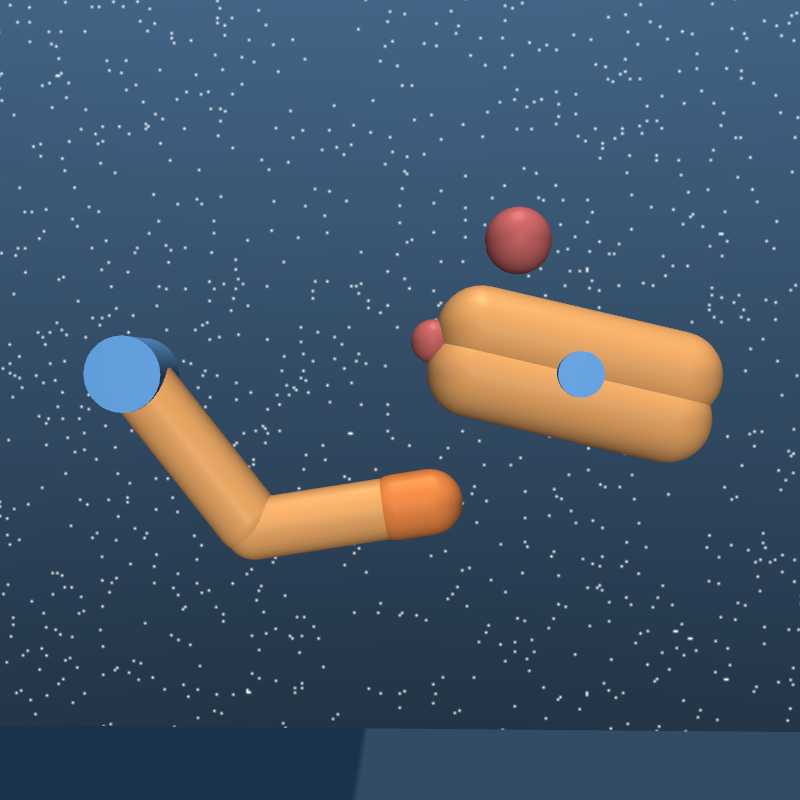}
\hspace{\myhsep}
\includegraphics[width=\mywidth]{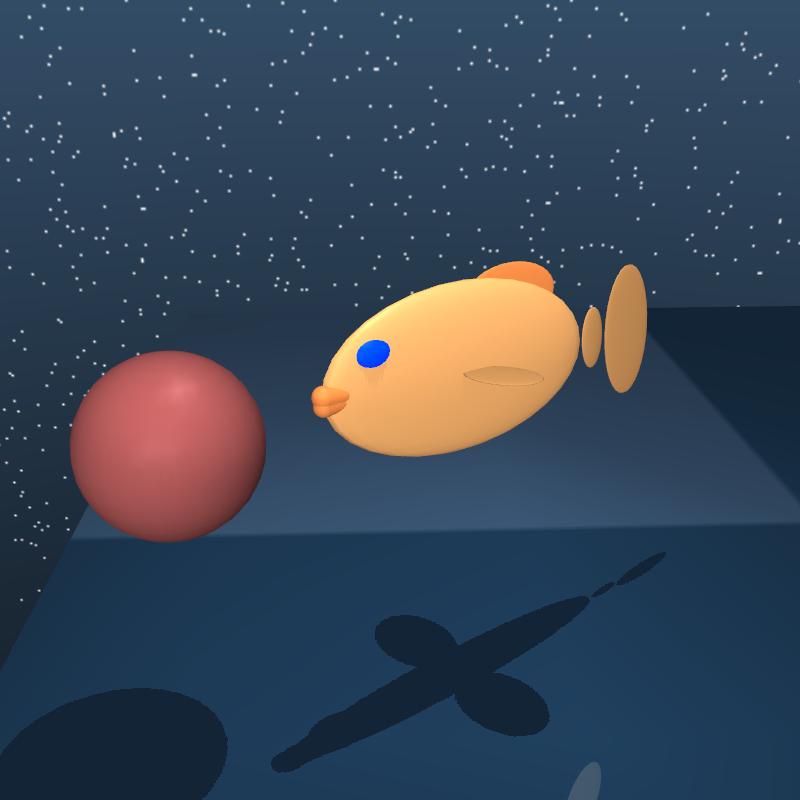}
\hspace{\myhsep}
\includegraphics[width=\mywidth]{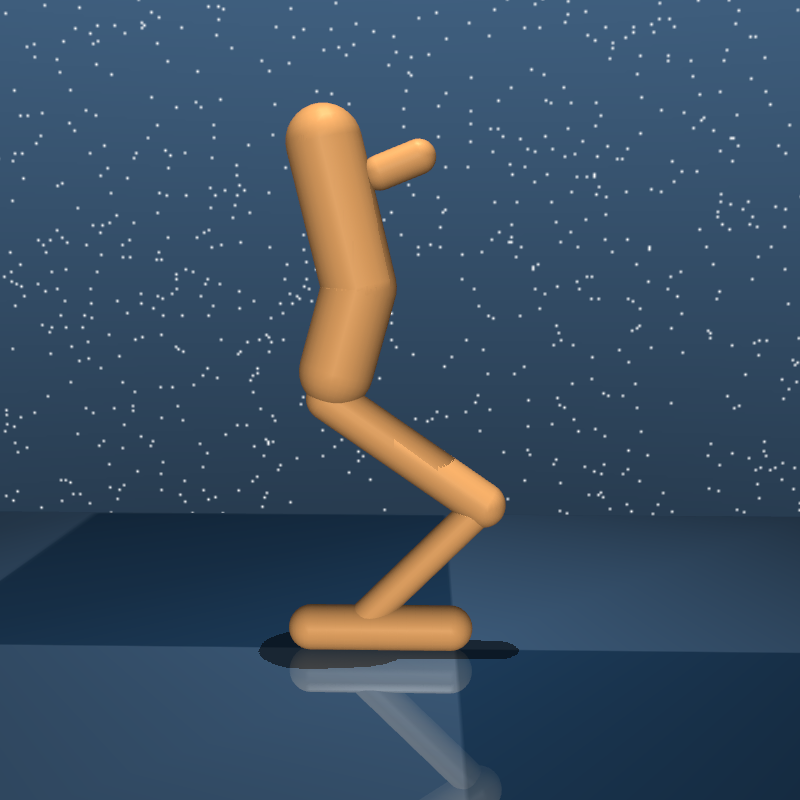}\\
\includegraphics[width=\mywidth]{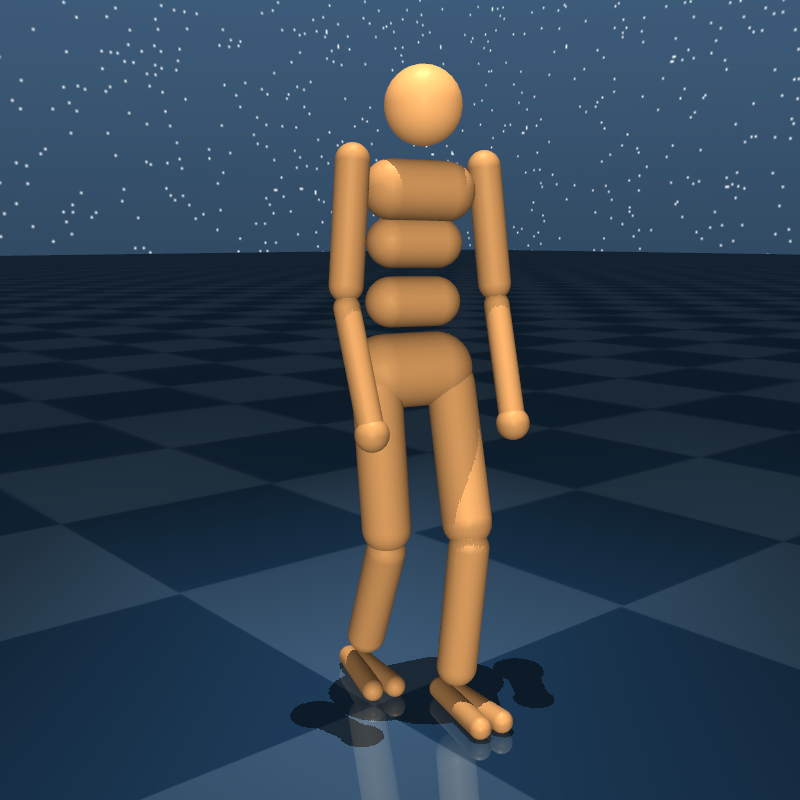}
\hspace{\myhsep}
\includegraphics[width=\mywidth]{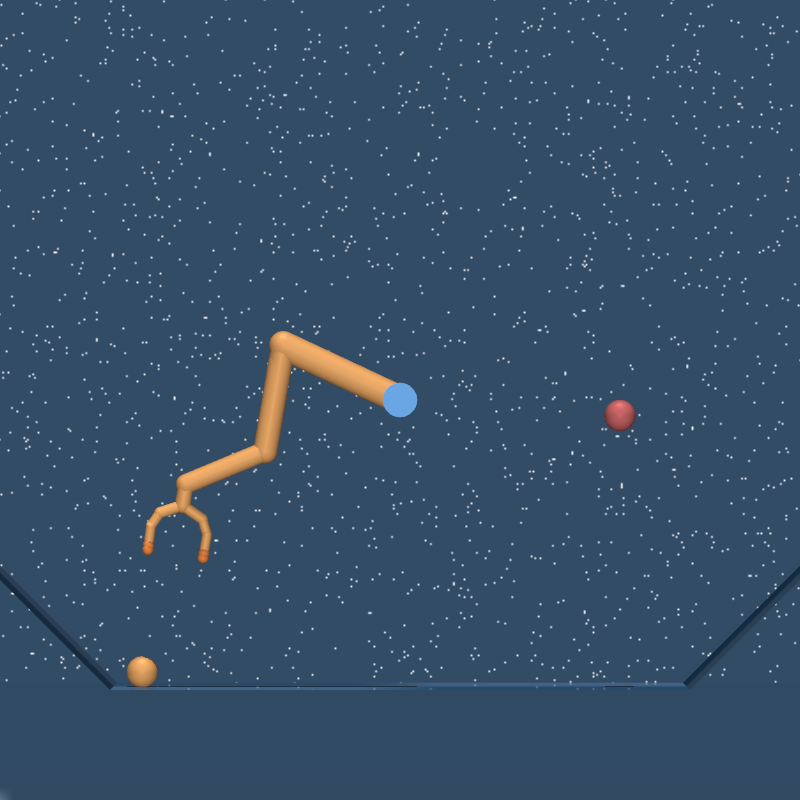}
\hspace{\myhsep}
\includegraphics[width=\mywidth]{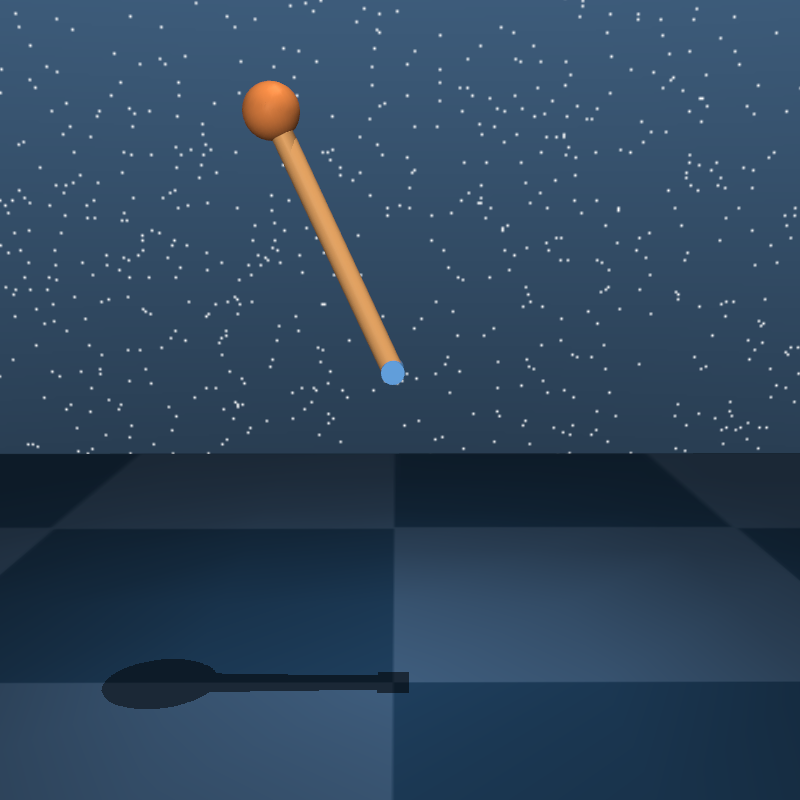}
\hspace{\myhsep}
\includegraphics[width=\mywidth]{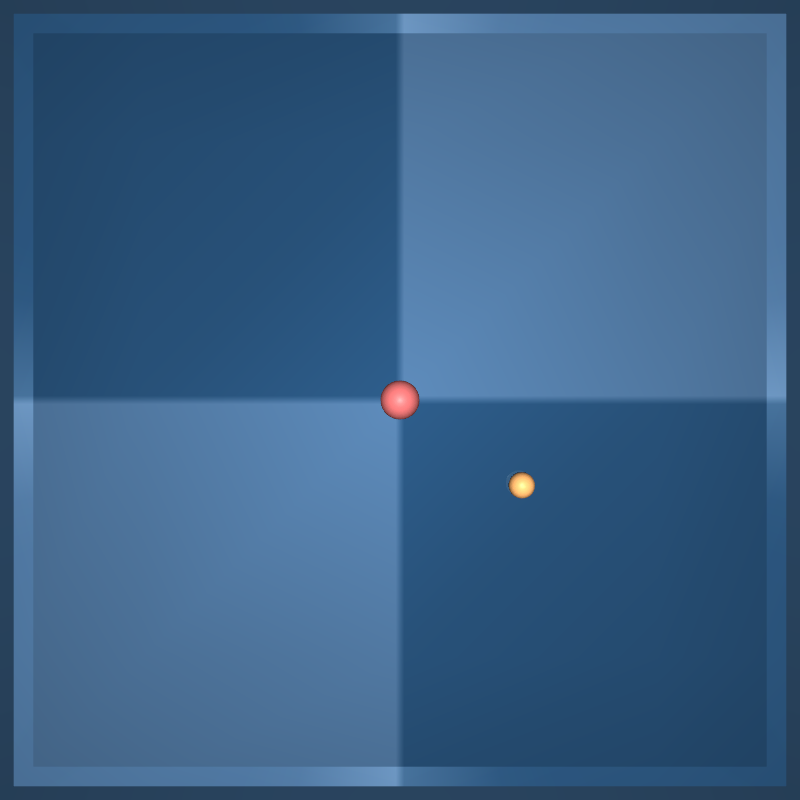}
\hspace{\myhsep}
\includegraphics[width=\mywidth]{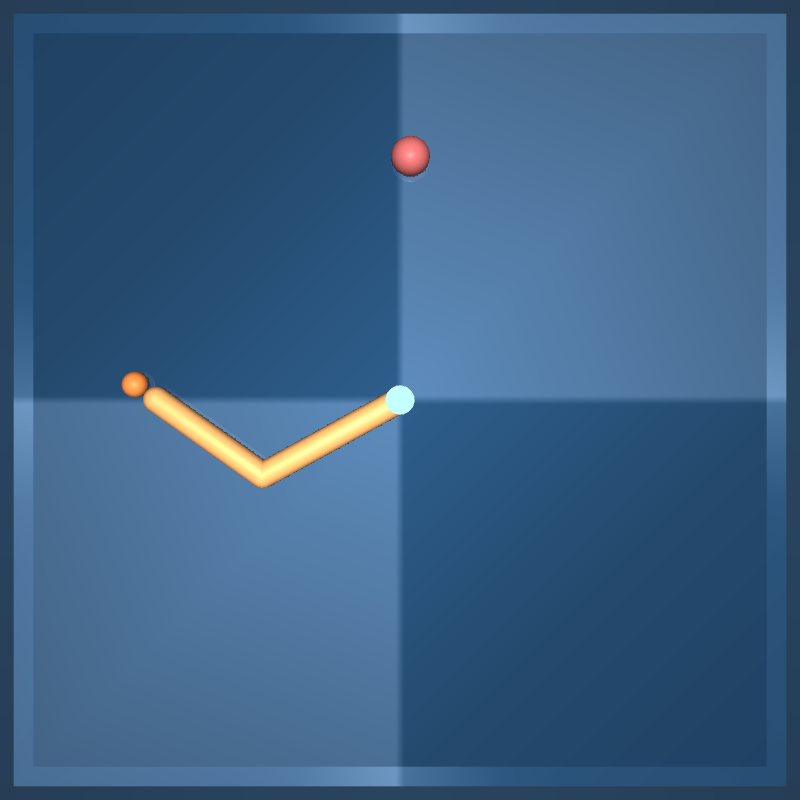}
\hspace{\myhsep}
\includegraphics[width=\mywidth]{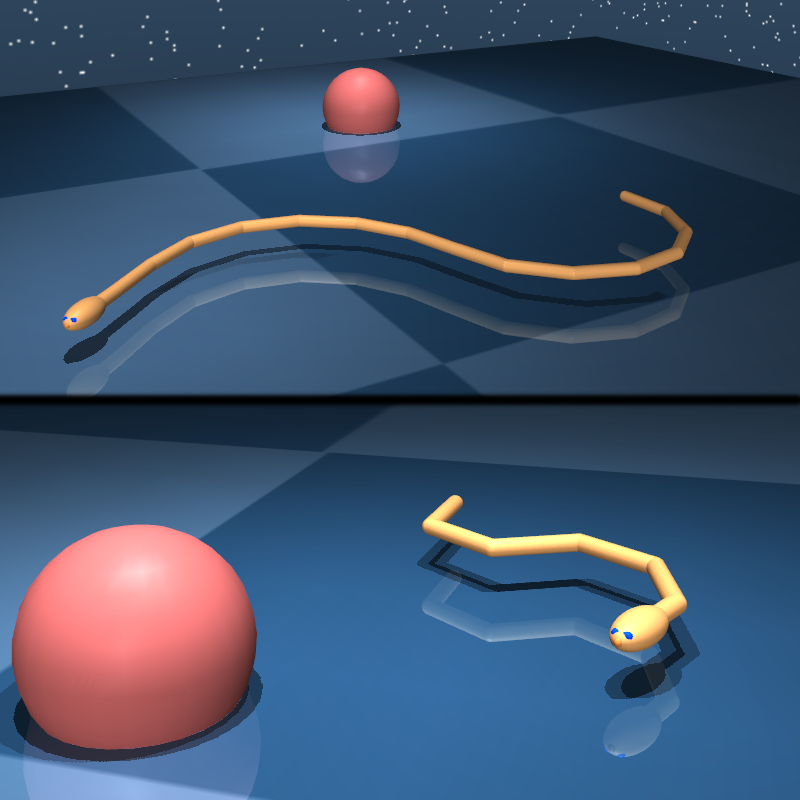}
\hspace{\myhsep}
\includegraphics[width=\mywidth]{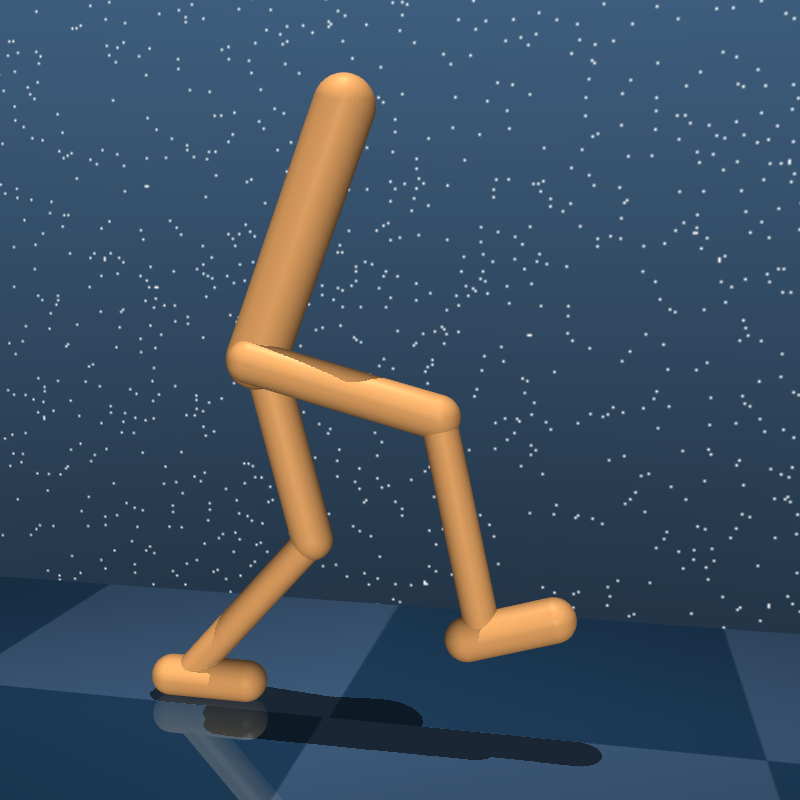}
\caption{\scriptsize Benchmarking domains. \textit{Top}: Acrobot, Ball-in-cup, Cart-pole, Cheetah, Finger, Fish, Hopper. \textit{Bottom}: Humanoid, Manipulator, Pendulum, Point-mass, Reacher, Swimmer (6 and 15 links), Walker.}
\label{fig:images}
\end{minipage}
\end{figure*}

\begin{abstract}
\noindent The \textit{DeepMind Control Suite} is a set of continuous control tasks with a standardised structure and interpretable rewards, intended to serve as performance benchmarks for reinforcement learning agents. The tasks are written in Python and powered by the MuJoCo physics engine, making them easy to use and modify. We include benchmarks for several learning algorithms. The \textit{Control Suite} is publicly available at \href{https://www.github.com/deepmind/dm_control}{\texttt{github.com/deepmind/dm\_control}}. A video summary of all tasks is available at \href{https://youtu.be/rAai4QzcYbs}{\texttt{youtu.be/rAai4QzcYbs}}.
\end{abstract}




\section{Introduction}
Controlling the physical world is an integral part and arguably a prerequisite of general intelligence. Indeed, the only known example of general-purpose intelligence emerged in primates which had been manipulating the world for millions of years.

Physical control tasks share many common properties and it is sensible to consider them as a distinct class of behavioural problems. Unlike board games, language and other symbolic domains, physical tasks are fundamentally \textit{continuous} in state, time and action. Their dynamics are subject to second-order equations of motion, implying that the underlying state is composed of position-like and velocity-like variables, while state derivatives are acceleration-like. Sensory signals (i.e.\ observations) usually carry meaningful physical units and vary over corresponding timescales. 

This decade has seen rapid progress in the application of Reinforcement Learning (RL) techniques to difficult problem domains such as video games \citep{Mnih2015}. The Arcade Learning Environment (ALE, \citealt{bellemare2012arcade}) was a vital facilitator of these developments, providing a set of standard benchmarks for evaluating and comparing learning algorithms. The \textit{DeepMind Control Suite} provides a similar set of standard benchmarks for continuous control problems. 



The OpenAI Gym~\citep{brockman2016gym} currently includes a set of continuous control domains that has become the de-facto benchmark in continuous RL~\citep{duan2016benchmarking, 2017deepRLmatters}. The Control Suite is also a set of tasks for benchmarking continuous RL algorithms, with a few notable differences. We focus exclusively on continuous control, e.g.\ separating observations with similar units (position, velocity, force etc.) rather than concatenating into one vector. Our unified reward structure (see below) offers interpretable learning curves and aggregated suite-wide performance measures. Furthermore, we emphasise high-quality well-documented code using uniform design patterns, offering a readable, transparent and easily extensible codebase.  Finally, the Control Suite has equivalent domains to all those in the Gym while adding many more\footnote{With the notable exception of Philipp Moritz's ``ant'' quadruped, which we intend to replace soon, see Future Work.}.


In Section 2 we explain the general structure of the Control Suite and in Section~3 we describe each domain in detail. In Sections 4 and 5 we document the high and low-level Python APIs, respectively. Section 6 is devoted to our benchmarking results. We then conclude and provide a roadmap for future development.

\section{Structure and Design}

The \textit{DeepMind Control Suite} is a set of stable, well-tested continuous control tasks that are easy to use and modify. Tasks are written in \href{https://www.python.org/}{Python} and physical models are defined using \href{http://mujoco.org/book/modeling.html}{MJCF}. Standardised action, observation and reward structures make benchmarking simple and learning curves easy to interpret. 

\subsubsection*{Model and Task verification}

Verification in this context means making sure that the physics simulation is stable and that the task is solvable: 

\begin{itemize}
\item Simulated physics can easily destabilise and diverge, mostly due to errors introduced by time discretisation. Smaller time-steps are more stable, but require more computation per unit simulation time, so the choice of time-step is always a trade-off between stability and speed~\citep{erez2015simulation}. What's more, learning agents are  better at discovering and exploiting instabilities.\footnote{This phenomenon, sometimes known as \textit{Sims' Law}, was first articulated in \citep{sims1994evolving}: ``Any bugs that allow energy leaks from non-conservation, or even round-off errors, will inevitably be discovered and exploited''.}

\item It is surprisingly easy to write tasks that are much easier or harder than intended, that are impossible to solve or that can be solved by very different strategies than expected (i.e.\ ``cheats''). To prevent these situations, the Atari{\scriptsize ™} games that make up ALE were extensively tested over more than 10 man-years\footnote{Marc Bellemare, personal communication.}. However, continuous control domains cannot be solved by humans, so a different approach must be taken.
\end{itemize}

In order to tackle both of these challenges, we ran variety of learning agents (e.g.\ \citealt{lillicrap2015continuous, mnih2016asynchronous}) against all tasks, and iterated on each task's design until we were satisfied that the physics was stable and non-exploitable, and that the task is solved correctly by at least one agent. Tasks that are solvable by some learning agent were collated into the \mono{benchmarking} set. Tasks were not solved by any learning agent are in the \mono{extra} set of tasks.

\subsection*{Reinforcement Learning}
A continuous Markov Decision Process (MDP) is given by a set of states $\mathcal{S}$, a set of actions $\mathcal{A}$, a dynamics (transition) function $\mathbf{f}(\mathbf{s},\mathbf{a})$, an observation function $\mathbf{o}(\mathbf{s}, \mathbf{a})$ and a scalar reward function $r(\mathbf{s}, \mathbf{a})$.

\begin{figure}[ht]
\includegraphics[width=0.95\textwidth]{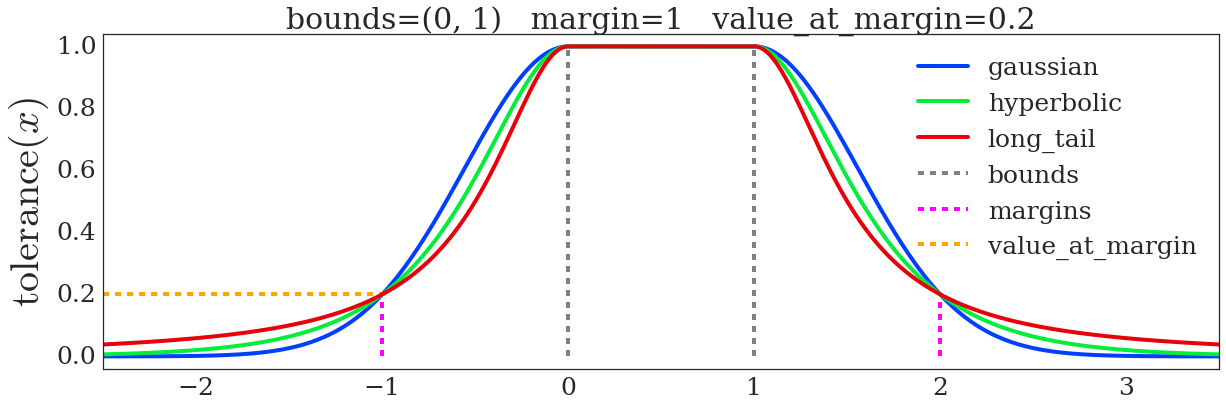}\\
\includegraphics[width=0.95\textwidth]{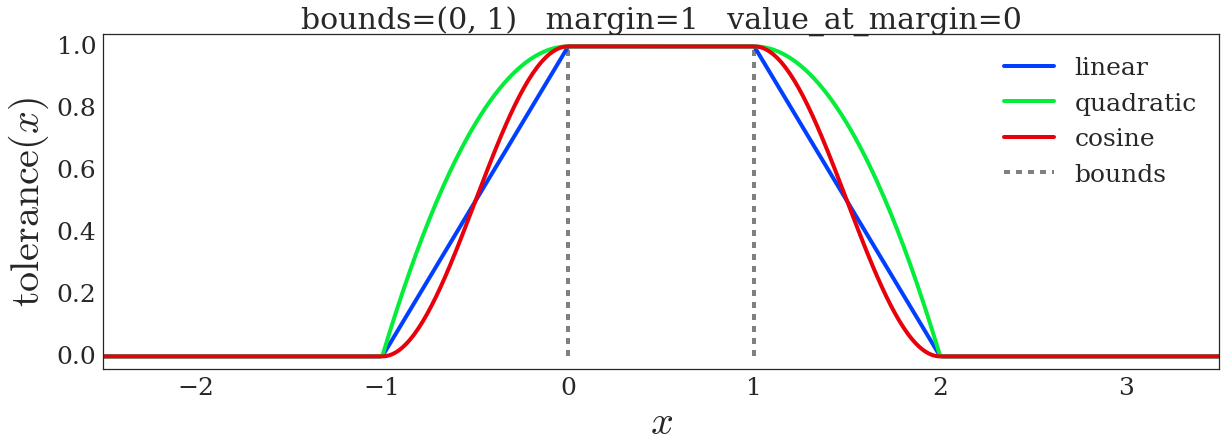}
\caption{The \mono{tolerance(x, bounds=(lower, upper))} function will return 1 if \mono{x} is within the \mono{bounds} interval and 0 otherwise. If the optional \mono{margin} argument is given, the output will decrease smoothly with distance from the interval, taking a value of \mono{value\_at\_margin} at a distance of \mono{margin}. Several types of sigmoid-like functions are available. \textbf{Top:} Three infinite-support sigmoids, for which \mono{value\_at\_margin} must be positive. \textbf{Bottom:} Three finite-support support sigmoids with \mono{value\_at\_margin=0}.}
\label{fig:tolerance}
\end{figure}

\begin{description}[leftmargin=1cm]
\item[State:]
The state $\mathbf{s}$ is a vector of real numbers $\mathcal{S} \equiv \mathbb{R}^{\dim(\mathcal{S})}$, with the exception of spatial orientations which are represented by unit quaternions $\in SU(2)$. States are initialised in some subset $\mathcal{S}_\textit{0}\subseteq \mathcal{S}$ by the \mono{begin\_episode()} method. To avoid memorised ``rote'' solutions $\mathcal{S}_\textit{0}$ is never a single state.

\item[Action:]
With the exception of the LQR domain (see below), the action vector is in the unit box $\mathbf{a}\in \mathcal{A} \equiv \left[-1, 1\right]^{\dim(\mathcal{A})}$.

\item[Dynamics:]
While the state notionally evolves according to a continuous ordinary differential equation $\dot{\mathbf{s}} = \mathbf{f}_c(\mathbf{s},\mathbf{a})$, in practice temporal integration is discrete\footnote{Most domains use MuJoCo's default semi-implicit Euler integrator, a few which have smooth, nearly energy-conserving dynamics use 4th-order Runge Kutta.} with some fixed, finite time-step: $\mathbf{s}_{t+h} = \mathbf{f}(\mathbf{s}_t,\mathbf{a}_t)$.

\item[Observation:]
The function $\mathbf{o}(\mathbf{s}, \mathbf{a})$ describes the observations available to the learning agent. With the exception of \mono{point-mass:hard} (see below), all tasks are strongly observable, i.e. the state can be recovered from a single observation. Observation features which depend only on the state (position and velocity) are functions of the current state. Features which are also dependent on controls (e.g. touch sensor readings) are functions of the previous transition. Observations are implemented as a Python \mono{OrderedDict}.

\item[Reward:]
The range of rewards in the Control Suite, with the exception of the LQR domain, are in the unit interval $r(\mathbf{s}, \mathbf{a}) \in [0, 1]$. Some tasks have ``sparse'' rewards $r(\mathbf{s}, \mathbf{a}) \in \{0,1\}$. This structure is facilitated by the \mono{tolerance()} function, see Figure \ref{fig:tolerance}. Since terms produced by \mono{tolerance()} are in the unit interval, both \emph{averaging} and \emph{multiplication} operations maintain that property, facillitating cost design.

\item[Termination and Discount:]
Control problems are classified as finite-horizon, first-exit and infinite-horizon~\citep{bertsekas1995dynamic}. Control Suite tasks have no terminal states or time limit and are therefore of the infinite-horizon variety. Notionally the objective is the continuous-time infinite-horizon average return $\lim_{T\rightarrow \infty} T^{-1}\int_{0}^{T}r(\mathbf{s}_t,\mathbf{a}_t)dt$, but in practice all of our agents internally use the discounted formulation $\int_{0}^{\infty}e^{-t/\tau}r(\mathbf{s}_t,\mathbf{a}_t)dt$ or, in discrete time
$\sum_{i=0}^\infty \gamma^i r(\mathbf{s}_i,\mathbf{a}_i)$,
where $\gamma=e^{-h/\tau}$ is the discount factor. In the limit $\tau \rightarrow \infty$ (equivalently $\gamma \rightarrow 1$), the policies of the discounted-horizon and average-return formulations are identical.

\item[Evaluation:]
While agents are expected to optimise for infinite-horizon returns, these are difficult to measure. As a proxy we use fixed-length episodes of 1000 time steps. Since all reward functions are designed so that $r \approx 1$ at or near a goal state, learning curves measuring total returns all have the same y-axis limits of $\mathbf{[0, 1000]}$, making them easier to interpret. 
\end{description}

\subsubsection*{MuJoCo physics}
MuJoCo~\citep{todorov2012mujoco} is a fast, minimal-coordinate, continuous-time physics engine. It compares favourably to other popular engines~\citep{erez2015simulation}, especially for articulated, low-to-medium degree-of-freedom (DoF) models in contact with other bodies. The convenient~\href{http://mujoco.org/book/modeling.html}{MJCF} definition format and reconfigurable computation pipeline have made MuJoCo popular\footnote{Along with the MultiBody branch of the~\href{http://bulletphysics.org/wordpress/}{Bullet} physics engine.} for robotics and reinforcement learning research (e.g.\ \citealt{schulman2015trust}).

\section{Domains and Tasks} \label{sec:domains}
A \textbf{domain} refers to a physical model, while a \textbf{task} refers to an instance of that model with a particular MDP structure. For example the difference between the \mono{swingup} and \mono{balance} tasks of the \mono{cartpole} domain is whether the pole is initialised pointing downwards or upwards, respectively. In some cases, e.g.\ when the model is procedurally generated, different tasks might have different physical properties. Tasks in the Control Suite are collated into tuples according predefined tags. In particular, tasks used for benchmarking are in the \mono{BENCHMARKING} tuple, while those not used for benchmarking (because they are particularly difficult, or because they don't conform to the standard structure) are in the \mono{EXTRA} tuple. All suite tasks are accessible via the \mono{ALL\_TASKS} tuple. In the domain descriptions below, names are followed by three integers specifying the dimensions of the state, control and observation spaces i.e.\ $\Bigl(\dim(\mathcal{S}),\dim(\mathcal{A}),\dim(\mathcal{O})\Bigr)$. 

\vspace{-.2cm}
\myfigure{
\textbf{Pendulum (2, 1, 3):} The classic inverted pendulum. The torque-limited actuator is $1/6_{\mathrm{th}}$ as strong as required to lift the mass from motionless horizontal, necessitating several swings to swing up and balance. The \mono{swingup} task has a simple sparse reward: 1 when the pole is within $30^\circ$ of the vertical and 0 otherwise.
}{./figures/pendulum.png}

\myfigure{
\textbf{Acrobot (4, 1, 6):} The underactuated double pendulum, torque applied to the second joint. The goal is to swing up and balance. Despite being low-dimensional, this is not an easy control problem. The physical model conforms to \citep{coulom2002reinforcement} rather than the earlier \citep{spong1995swing}. Both \mono{swingup} and \mono{swingup\_sparse} tasks with smooth and sparse rewards, respectively.
}{./figures/acrobot.png}

\myfigure{
\textbf{Cart-pole (4, 1, 5):} Swing up and balance an unactuated pole by applying forces to a cart at its base. The physical model conforms to \citep{barto1983neuronlike}. Four benchmarking tasks: in \mono{swingup} and \mono{swingup\_sparse} the pole starts pointing down while in \mono{balance} and \mono{balance\_sparse} the pole starts near the upright. }{./figures/cart-pole.png}

\begin{figure}[H]
\floatbox[{\capbeside\thisfloatsetup{capbesideposition={right,top},
capbesidewidth=0.585\textwidth}}]{figure}[\FBwidth]
{\caption*{
\textbf{Cart-k-pole ($\mathbf{2k\!+\!2,\; 1,\; 3k\!+\!2}$):}
The cart-pole domain allows to procedurally adding more poles, connected serially.
Two non-benchmarking tasks, \mono{two\_poles} and \mono{three\_poles} are available.
}}
{\includegraphics[width=2.5cm]{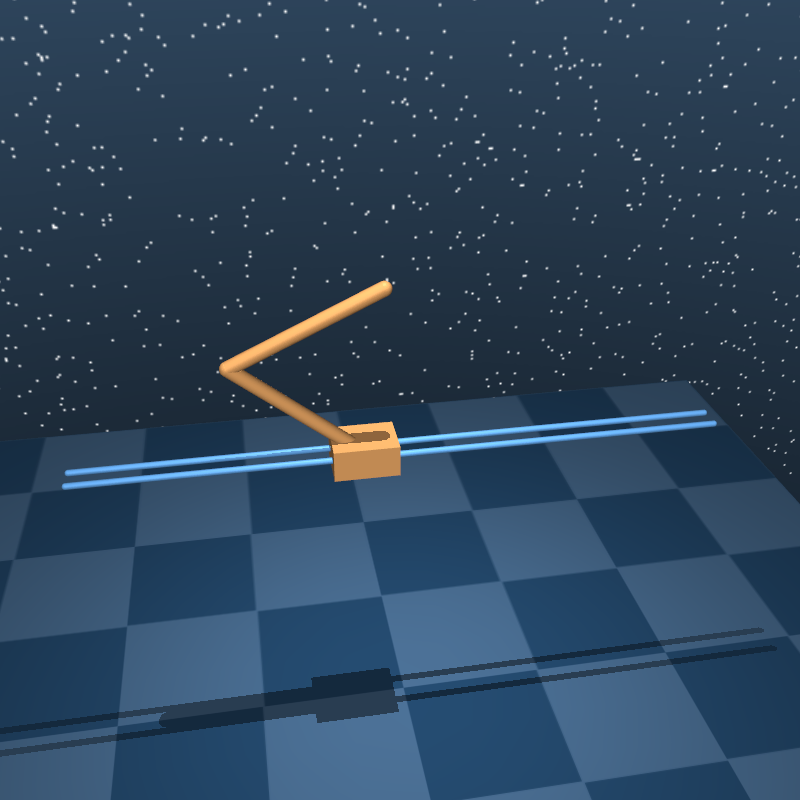}
\hspace{.1cm}
\includegraphics[width=2.5cm]{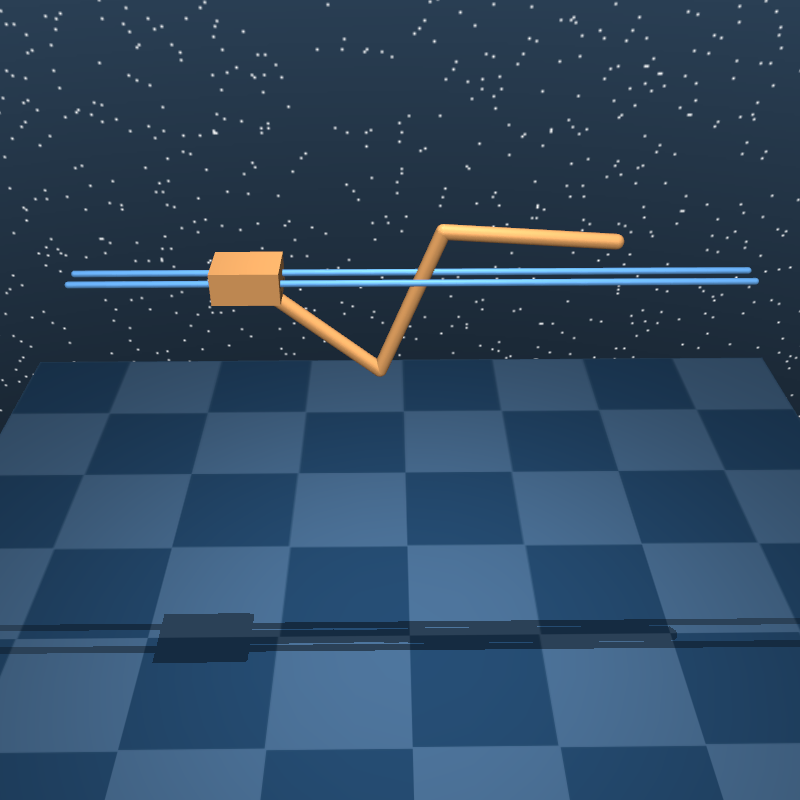}}
\end{figure}
\vspace{-.5cm}

\myfigure{
\textbf{Ball in cup (8, 2, 8):} A planar ball-in-cup task. An actuated planar receptacle can translate in the vertical plane in order to swing and catch a ball attached to its bottom. The \mono{catch} task has a sparse reward: 1 when the ball is in the cup, 0 otherwise. 
}{./figures/ball-in-cup.png}

\myfigure{
\textbf{Point-mass (4, 2, 4):} A planar point-mass receives a reward of 1 when within a target at the origin. In the \mono{easy} task, one of simplest in the suite, the 2 actuators correspond to the global $x$ and $y$ axes. In the \mono{hard} task the gain matrix from the controls to the axes is randomised for each episode, making it impossible to solve by memory-less agents; this task is not in the \mono{benchmarking} set. 
}{./figures/point-mass.png}

\myfigure{
\textbf{Reacher (4, 2, 7):} The simple two-link planar reacher with a randomised target location. The reward is one when the end effector penetrates the target sphere. In the \mono{easy} task the target sphere is bigger than on the \mono{hard} task (shown on the left).  
}{./figures/reacher.png}

\myfigure{
\textbf{Finger (6, 2, 12):} 
A 3-DoF toy manipulation problem based on \citep{tassa2010stochastic}. A planar `finger' is required to rotate a body on an unactuated hinge. In the \mono{turn\_easy} and \mono{turn\_hard} tasks, the tip of the free body must overlap with a target (the target is smaller for the \mono{turn\_hard} task). In the \mono{spin} task, the body must be continually rotated.
}{./figures/finger.png}

\myfigure{
\textbf{Hopper (14, 4, 15):} The planar one-legged hopper introduced in \citep{lillicrap2015continuous}, initialised in a random configuration. In the \mono{stand} task it is rewarded for bringing its torso to a minimal height. In the \mono{hop} task it is rewarded for torso height and forward velocity. 
}{./figures/hopper.png}

\myfigure{
\textbf{Fish (26, 5, 24):} A fish is required to swim to a target. This domain relies on MuJoCo's simplified fluid dynamics. Two tasks: in the \mono{upright} task, the fish is rewarded only for righting itself with respect to the vertical, while in the \mono{swim} task it is also rewarded for swimming to the target.
}{./figures/fish.png}

\myfigure{
\textbf{Cheetah (18, 6, 17):} A running planar biped based on \citep{wawrzynski2009real}.
The reward $r$ is linearly proportional to the forward velocity $v$ up to a maximum of $10_{\mathrm{m/s}}$ i.e.\ $r(v) = \max\bigl(0, \min(v/10, 1)\bigr)$.
}{./figures/cheetah.png}

\myfigure{
\textbf{Walker (18, 6, 24):} An improved planar walker based on the one introduced in \citep{lillicrap2015continuous}. In the \mono{stand} task reward is a combination of terms encouraging an upright torso and some minimal torso height. The \mono{walk} and \mono{run} tasks include a component encouraging forward velocity.
}{./figures/walker.png}

\myfigure{
\textbf{Manipulator (22, 5, 37):} A planar manipulator is rewarded for bringing an object to a target location. In order to assist with exploration, in \%10 of episodes the object is initialised in the gripper or at the target. Four \mono{manipulator} tasks: \mono{\{bring,insert\}\_\{ball,peg\}} of which only \mono{bring\_ball} is in the \mono{benchmarking} set. The other three are shown below.

}{./figures/manipulator.png}

\begin{figure}[H]
\floatbox[{\capbeside\thisfloatsetup{capbesideposition={right,top},
capbesidewidth=0.415\textwidth}}]{figure}[\FBwidth]
{\caption*{
\textbf{Manipulator extra:}
\mono{insert\_ball}: place the ball in the basket. \mono{bring\_peg}: bring the peg to the target peg (matching orientation). \mono{insert\_peg}: insert the peg into the slot.
}}
{\includegraphics[width=2.5cm]{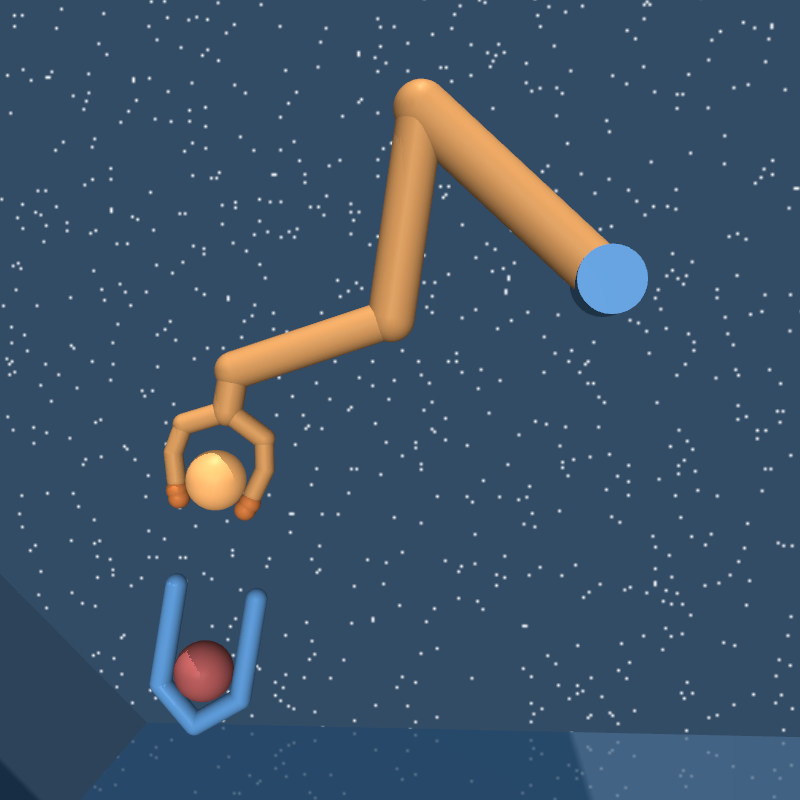}
\includegraphics[width=2.5cm]{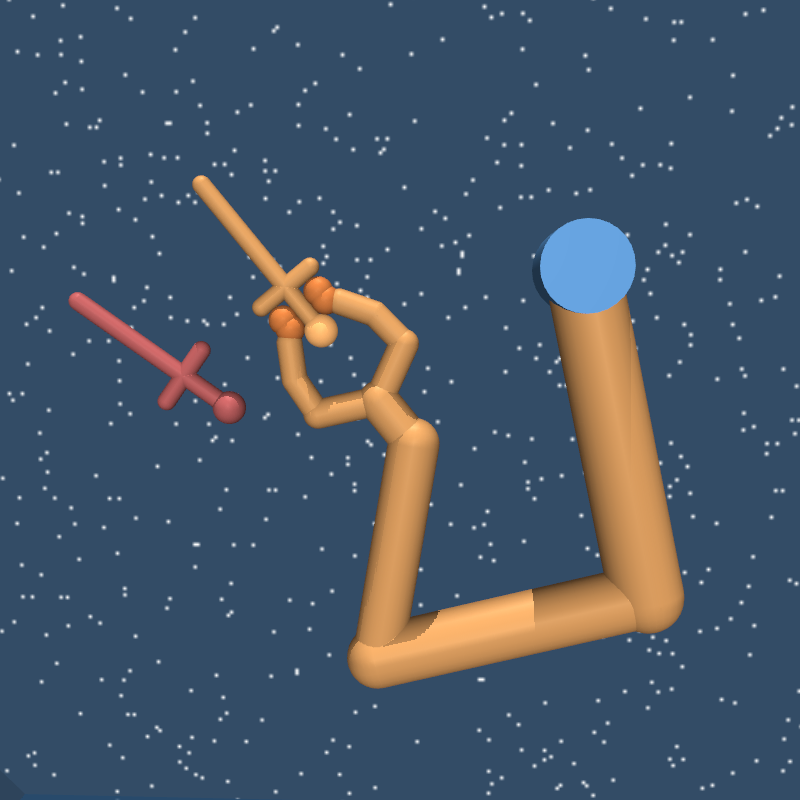}
\includegraphics[width=2.5cm]{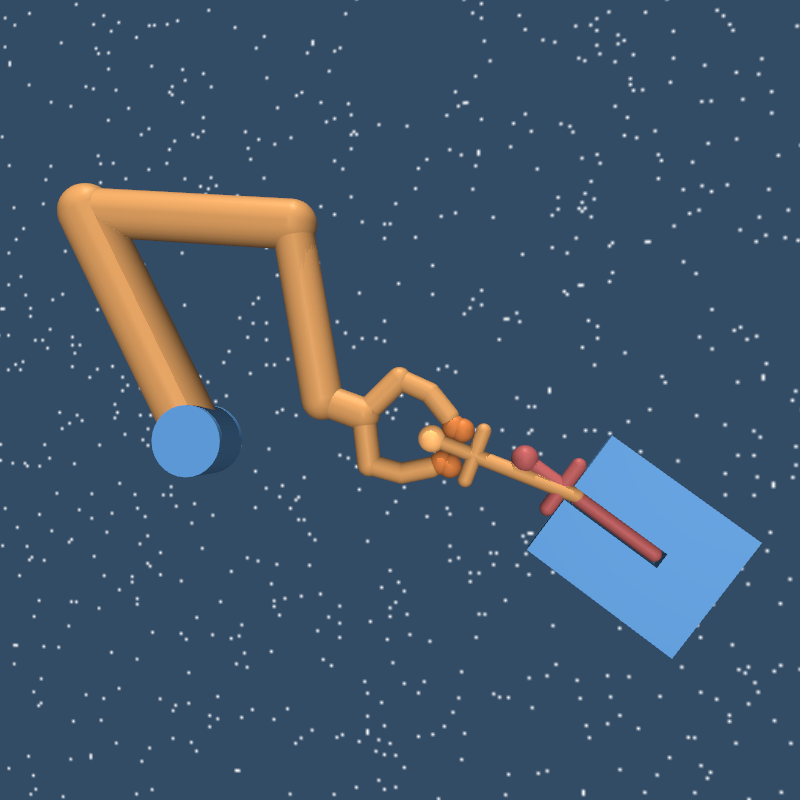}}
\end{figure}
\vspace{-.5cm}

\myfigure{
\textbf{Stacker ($\mathbf{6k\!+\!16,\; 5,\; 11k\!+\!26}$):} Stack $k$ boxes. Reward is given when a box is at the target and the gripper is away from the target, making stacking necessary. The height of the target is sampled uniformly from $\{1,\ldots,k\}$.
}{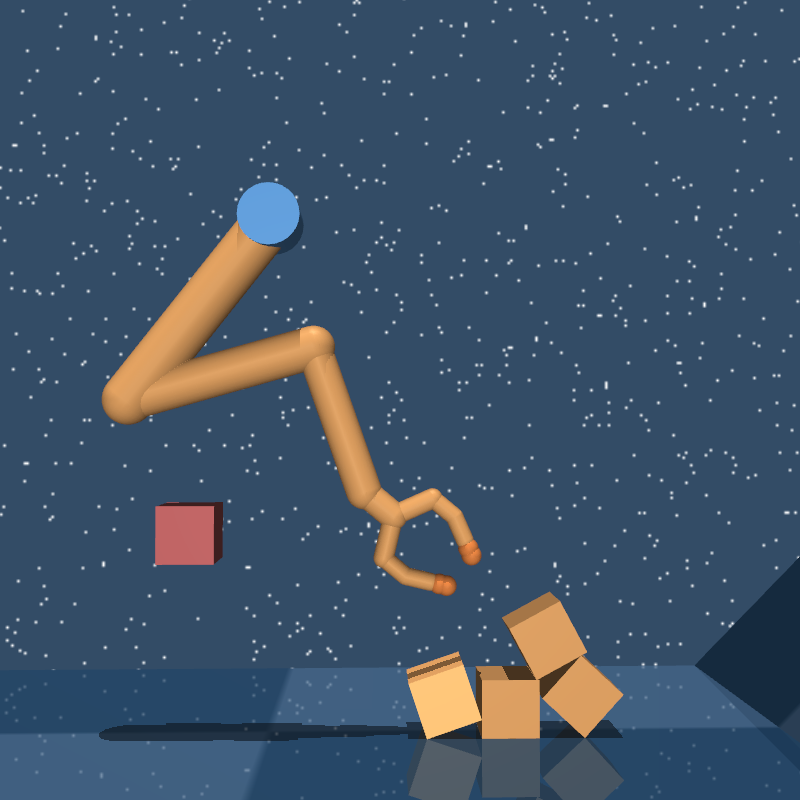}

\myfigure{
\textbf{Swimmer ($\mathbf{2k\!+\!4,\; k\!-\!1,\; 4k\!+\!1}$):}
This procedurally generated $k$-link planar swimmer is based on \citep{coulom2002reinforcement} but using MuJoCo's high-Reynolds fluid drag model. A reward of 1 is provided when the nose is inside the target and decreases smoothly with distance like a Lorentzian. The two instantiations provided in the \mono{benchmarking} set are the 6-link and 15-link swimmers.
}{./figures/swimmers.png}

\myfigure{
\textbf{Humanoid (54, 21, 67):} A simplified humanoid with 21 joints, based on the model in~\citep{tassa2012synthesis}. Three tasks: \mono{stand}, \mono{walk} and \mono{run} are differentiated by the desired horizontal speed of 0, 1 and $10_{m/s}$, respectively. Observations are in an egocentric frame and many movement styles are possible solutions e.g. running backwards or sideways. This facilitates exploration of local optima.
}{./figures/humanoid.png}

\myfigure{
\textbf{Humanoid\_CMU (124, 56, 137):} A humanoid body with 56 joints, adapted from \citep{merel2017learning} and based on the ASF model of subject~\#8 in the \citetalias{cmu_mocap}. This domain has the same \mono{stand}, \mono{walk} and \mono{run} tasks as the simpler humanoid. We include tools for parsing and playback of the CMU MoCap data, see below.
}{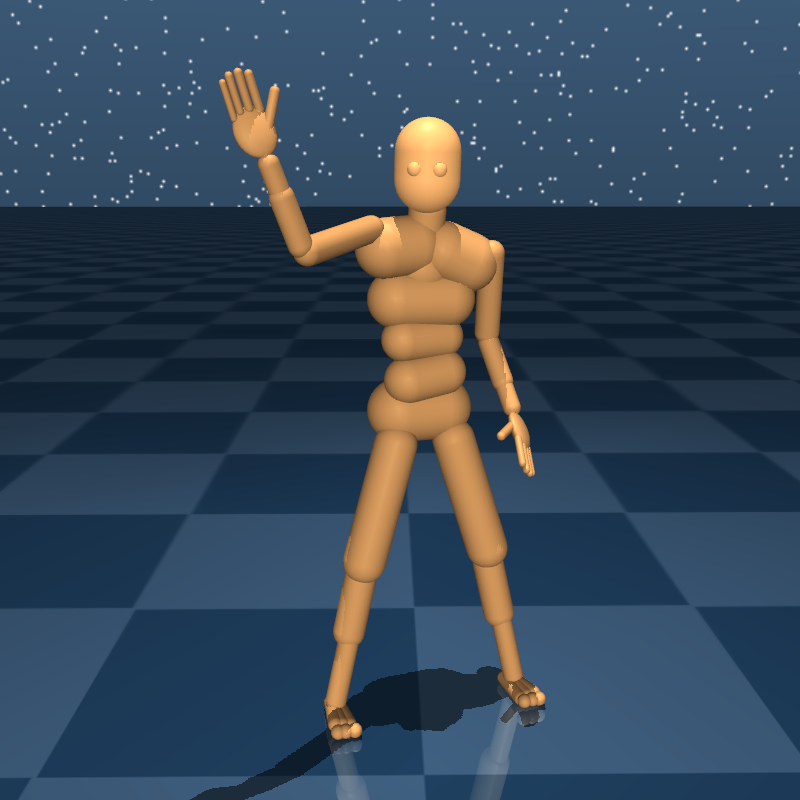}

\myfigure{
\textbf{LQR (2n, m, 2n):} $n$ masses, of which $m\leq n$ are actuated, move on linear joints which are connected serially. The reward is a quadratic in the position and controls. Analytic transition and control-gain matrices are extracted from MuJoCo and the optimal policy and value functions are computed in \mono{lqr\_solver.py} using Riccati iterations.
Since both controls and reward are unbounded, \mono{LQR} is not in the \mono{benchmarking} set. 
}{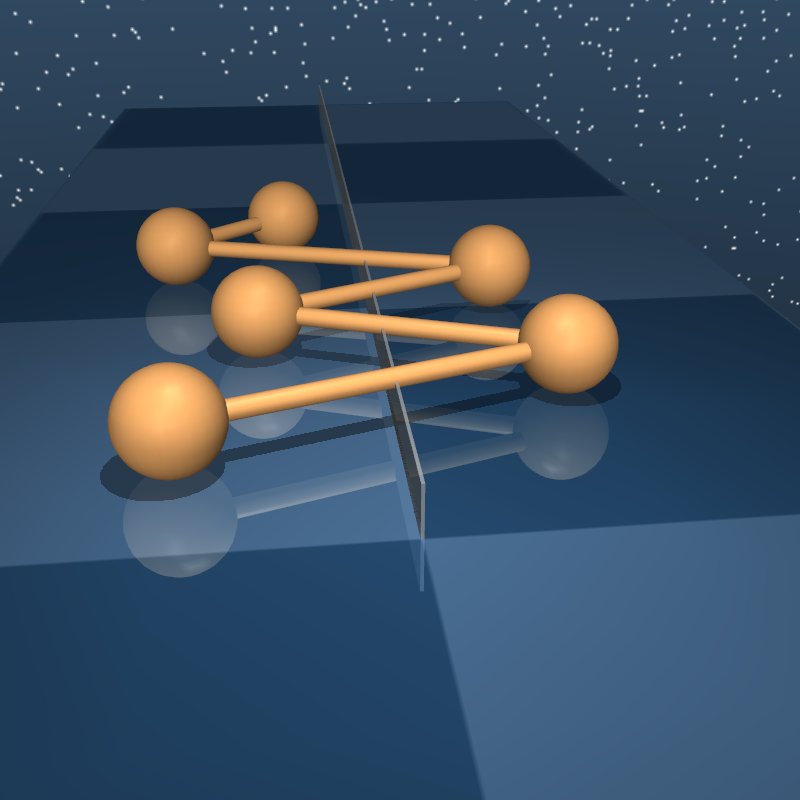}

\subsubsection*{CMU Motion Capture Data}
We enable \mono{humanoid\_CMU} to be used for imitation learning as in \cite{merel2017learning} by providing tools for parsing, conversion and playback of human motion capture data from the \citetalias{cmu_mocap}. The \mono{convert()} function in the \mono{parse\_amc} module loads an AMC data file and returns a sequence of configurations for the \mono{humanoid\_CMU} model. The example script \mono{CMU\_mocap\_demo.py} uses this function to generate a video.

\section{Reinforcement learning API}

In this section we describe the following Python code:
\begin{itemize}
    \item The \mono{environment.Base} class that defines generic RL interface.
    \item The \mono{suite} module that contains the domains and tasks defined in Section \ref{sec:domains}
    \item The underlying MuJoCo bindings and the \mono{mujoco.Physics} class that provides most of the functionality needed to interact with an instantiated MJCF model.
\end{itemize}

\subsection*{The RL Environment class}
The class \mono{environment.Base}, found within the \mono{dm\_control.rl.environment} module, defines the following abstract methods:
\begin{description}[leftmargin=0cm,itemindent=0.2cm,font=$\bullet$~]

\item[\mono{action\_spec()} \textnormal{and} \mono{observation\_spec()}] describe the actions accepted and the observations returned by an \mono{Environment}.
For all the tasks in the suite, actions are given as a single NumPy array. \mono{action\_spec()} returns an \mono{ArraySpec}, with attributes describing the shape, data type, and optional minimum and maximum bounds for the action arrays. Observations consist of an \mono{OrderedDict} containing one or more NumPy arrays. \mono{observation\_spec()} returns an \mono{OrderedDict} of \mono{ArraySpec}s describing the shape and data type of each corresponding observation.

\item[\mono{reset()} \textnormal{and} \mono{step()}] respectively start a new episode, and advance time given an action.

\end{description}
Starting an episode and running it to completion might look like
\begin{lstlisting}[language=Python]
spec = env.action_spec()
time_step = env.reset()
while not time_step.last():
  action = np.random.uniform(spec.minimum, spec.maximum, spec.shape)
  time_step = env.step(action)
\end{lstlisting}
Both \mono{reset()} and \mono{step()} return a \mono{TimeStep}  \mono{namedtuple} with fields \mono{[step\_type, reward, discount, observation]}:
\begin{description}[leftmargin=0cm,itemindent=0.2cm,font=$\bullet$~]
\item[\mono{step\_type}] is an enum taking a value in \mono{[FIRST, MID, LAST]}. The convenience methods \mono{first()}, \mono{mid()} and \mono{last()} return boolean values indicating whether the \mono{TimeStep}'s type is of the respective value.
\item[\mono{reward}] is a scalar float.
\item[\mono{discount}] is a scalar float $\gamma \in [0, 1]$.
\item[\mono{observation}] is an \mono{OrderedDict} of NumPy arrays matching the specification returned by \mono{observation\_spec()}.
\end{description}
Whereas the \mono{step\_type} specifies whether or not the episode is terminating, it is the \mono{discount} $\gamma$ that determines the termination type. $\gamma=0$ corresponds to a terminal state\footnote{i.e.\ where the sum of future reward is equal to the current reward.} as in the first-exit or finite-horizon formulations. A terminal \mono{TimeStep} with $\gamma=1$ corresponds to the infinite-horizon formulation. In this case an agent interacting with the environment should treat the episode as if it could have continued indefinitely, even though the sequence of observations and rewards is truncated.
All Control Suite tasks with the exception of \mono{LQR}\footnote{The \mono{LQR} task terminates with $\gamma=0$ when the state is very close to 0, which is a proxy for the infinite exponential convergence of stabilised linear systems.} return $\gamma=1$ at every step, including on termination.

\subsection*{The \mono{suite} module}

To load an environment representing a task from the suite, use \mono{suite.load()}:
\begin{lstlisting}[language=Python]
from dm_control import suite

# Load one task:
env = suite.load(domain_name="cartpole", task_name="swingup")

# Iterate over a task set:
for domain_name, task_name in suite.BENCHMARKING:
  env = suite.load(domain_name, task_name)
  ...
\end{lstlisting}
Wrappers can be used to modify the behaviour of control environments:

\begin{description}[leftmargin=.3cm,itemindent=-.3cm]

\item{\textbf{Pixel observations}}

By default, Control Suite environments return low-dimensional feature observations. The \mono{pixel.Wrapper} adds or replaces these with images.
\begin{lstlisting}[language=Python]
from dm_control.suite.wrappers import pixels
env = suite.load("cartpole", "swingup")
env_and_pixels = pixels.Wrapper(env)
# Replace existing features by pixel observations.
env_only_pixels = pixels.Wrapper(env, pixel_only=False)
# Pixel observations in addition to existing features.
\end{lstlisting}

\item{\textbf{Reward visualisation}}

Models in the Control Suite use a common set of colours and textures for visual uniformity. As illustrated in the \href{https://youtu.be/rAai4QzcYbs}{video}, this also allows us to modify colours in proportion to the reward, providing a convenient visual cue. 
\begin{lstlisting}[language=Python]
env = suite.load("fish", "swim", task_kwargs, visualize_reward=True)
\end{lstlisting}

\end{description}

\section{MuJoCo Python interface}
While the \mono{environment.Base} class is specific to the Reinforcement Learning scenario, the underlying bindings and \mono{mujoco.Physics} class provide a general-purpose wrapper of the MuJoCo engine.
We use Python's \href{https://docs.python.org/3/library/ctypes.html}{\texttt{ctypes}} library to bind to MuJoCo structs, enums and functions.

\begin{description}[leftmargin=.3cm,itemindent=-.3cm]

\item{\textbf{Functions}}

The bindings provide easy access to all MuJoCo library functions, automatically converting NumPy arrays to data pointers where appropriate.
\begin{lstlisting}[language=Python]
from dm_control.mujoco.wrapper.mjbindings import mjlib
import numpy as np

quat = np.array((.5, .5, .5, .5))
mat = np.zeros((9))
mjlib.mju_quat2Mat(mat, quat)

print("MuJoCo can convert this quaternion:")
print(quat)
print("To this rotation matrix:")
print(mat.reshape(3,3))
\end{lstlisting}
\vspace{-.2cm}
\begin{lstlisting}[backgroundcolor=\color{palegray}]
MuJoCo can convert this quaternion:
[ 0.5  0.5  0.5  0.5]
To this rotation matrix:
[[ 0.  0.  1.]
 [ 1.  0.  0.]
 [ 0.  1.  0.]]
\end{lstlisting}

\item{\textbf{Enums}}
\begin{lstlisting}[language=Python]
from dm_control.mujoco.wrapper.mjbindings import enums
print(enums.mjtJoint)
\end{lstlisting}
\vspace{-.2cm}
\begin{lstlisting}[backgroundcolor=\color{palegray}]
mjtJoint(mjJNT_FREE=0, mjJNT_BALL=1, mjJNT_SLIDE=2, mjJNT_HINGE=3)
\end{lstlisting}

\end{description}

\subsection*{The \mono{Physics} class}
The \mono{Physics} class encapsulates MuJoCo's most commonly used  functionality.
\begin{description}[leftmargin=.3cm,itemindent=-.3cm]
\item{\textbf{Loading an MJCF model}}

The \mono{Physics.from\_xml\_string()} method loads an MJCF model and returns a \mono{Physics} instance: 
\begin{lstlisting}[language=Python]
simple_MJCF = """
<mujoco>
  <worldbody>
    <light name="top" pos="0 0 1.5"/>
    <geom name="floor" type="plane" size="1 1 .1"/>
    <body name="box" pos="0 0 .3">
      <joint name="up_down" type="slide" axis="0 0 1"/>
      <geom name="box" type="box" size=".2 .2 .2" rgba="1 0 0 1"/>
      <geom name="sphere" pos=".2 .2 .2" size=".1" rgba="0 1 0 1"/>
    </body>
  </worldbody>
</mujoco>
"""
physics = mujoco.Physics.from_xml_string(simple_MJCF)
\end{lstlisting}

\item{\textbf{Rendering}}

The \mono{Physics.render()} method outputs a numpy array of pixel values. 
\begin{lstlisting}[language=Python]
pixels = physics.render()
\end{lstlisting}
\includegraphics[width=4cm]{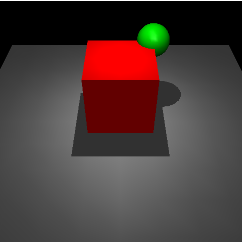}

Optional arguments to \mono{render} can be used to specify the resolution, camera ID and whether to render RGB or depth images. 
\item{\textbf{\mono{Physics.model} and \mono{Physics.data}}}

MuJoCo's \mono{mjModel} and \mono{mjData} structs, describing static and dynamic simulation parameters, can be accessed via the \mono{model} and \mono{data} properties of \mono{Physics}. They contain NumPy arrays that have direct, writeable views onto MuJoCo's internal memory. Because the memory is owned by MuJoCo an attempt to overwrite an entire array will fail:
\begin{lstlisting}[language=Python]
# This will fail:
physics.data.qpos = np.random.randn(physics.model.nq)
# This will succeed:
physics.data.qpos[:] = np.random.randn(physics.model.nq)
\end{lstlisting}

\item{\textbf{Setting the state with \mono{reset\_context()}}}

When setting the MujoCo state, derived quantities like global positions or sensor measurements are not updated. In order to facilitate synchronisation of derived quantities we provide the \mono{Physics.reset\_context()} context:
\begin{lstlisting}[language=Python]
with physics.reset_context():
   # mj_reset() is called upon entering the context.
   physics.data.qpos[:] = ...  # Set position,
   physics.data.qvel[:] = ...  # velocity
   physics.data.ctrl[:] = ...  # and control.
# mj_forward() is called upon exiting the context. Now all derived
# quantities and sensor measurements are up-to-date.
\end{lstlisting}
\item{\textbf{Running the simulation}}

The \mono{physics.step()} method is used to advance the simulation. Note that this method does not directly call MuJoCo's \mono{mj\_step()} function. 
At the end of an \mono{mj\_step} the state is updated, but the intermediate quantities stored in \mono{mjData} were computed with respect to the \emph{previous} state.
To keep these derived quantities as closely synchronised with the current simulation state as possible, we use the fact that MuJoCo partitions \mono{mj\_step} into two parts: \mono{mj\_step1}, which depends only on the state and \mono{mj\_step2}, which also depends on the control. Our \mono{physics.step} first executes \mono{mj\_step2} (assuming \mono{mj\_step1} has already been called), and then calls \mono{mj\_step1}, beginning the next step\footnote{In the case of Runge-Kutta integration, we simply conclude each RK4 step with an \mono{mj\_step1}.}. The upshot is that quantities that depend only on position and velocity (e.g.\ camera pixels) are synchronised with the current state, while quantities that depend on force/acceleration (e.g.\ touch sensors) are with respect to the previous transition.

\item{\textbf{Named indexing}}

It is often more convenient and less error-prone to refer to elements in the simulation by name rather than by index. \mono{Physics.named.model} and \mono{Physics.named.data} provide array-like containers that provide convenient named views:
\begin{lstlisting}[language=Python]
print("The geom_xpos array:")
print(physics.data.geom_xpos)
print("Is much easier to inspect using Physics.named")
print(physics.named.data.geom_xpos)
\end{lstlisting}
\begin{lstlisting}[backgroundcolor=\color{palegray}]
The data.geom_xpos array:
[[ 0.   0.   0. ]
 [ 0.   0.   0.3]
 [ 0.2  0.2  0.5]]
Is much easier to inspect using Physics.named:
           x         y         z         
0  floor [ 0         0         0       ]
1    box [ 0         0         0.3     ]
2 sphere [ 0.2       0.2       0.5     ]
\end{lstlisting}
These containers can be indexed by name for both reading and writing, and support most forms of NumPy indexing:
\begin{lstlisting}[language=Python]
with physics.reset_context():
  physics.named.data.qpos["up_down"] = 0.1
print(physics.named.data.geom_xpos["box", ["x", "z"]])
\end{lstlisting}
\begin{lstlisting}[backgroundcolor=\color{palegray}]
[ 0.   0.4]
\end{lstlisting}
Note that in the example above we use a joint name to index into the generalised position array \mono{qpos}. Indexing into a multi-DoF \mono{ball} or \mono{free} joint would output the appropriate slice.

\noindent We also provide convenient access to MuJoCo's \mono{mj\_id2name} and \mono{mj\_name2id}:
\begin{lstlisting}[language=Python]
physics.model.id2name(0, "geom")
\end{lstlisting}
\begin{lstlisting}[backgroundcolor=\color{palegray},upquote=true]
'floor'
\end{lstlisting}

\end{description}

\section{Benchmarking}

We provide baselines for two commonly employed deep reinforcement learning algorithms A3C~\citep{williams1991function,mnih2016asynchronous} and DDPG~\citep{lillicrap2015continuous}, as well as the recently introduced D4PG~\citep{d4pg}. We refer to the relevant papers for algorithm motivation and details and here provide only hyperparameter, network architecture, and training configuration information (see relevant sections below).

We study both the case of learning with state derived features as observations and learning from raw-pixel inputs for all the tasks in the Control Suite.  It is of course possible to look at control via combined state features and pixel features, but we do not study this case here.  We present results for both final performance and learning curves that demonstrate aspects of data-efficiency and stability of training. 

Establishing baselines for reinforcement learning problems and algorithms is notoriously difficult~\citep{islam2017reproducibility,2017deepRLmatters}.  Though we describe results for well-functioning implementations of the algorithms we present, it may be possible to perform better on these tasks with the same algorithms. For a given algorithm we ran experiments with a similar network architecture, set of hyperparameters, and training configuration as described in the original papers.  We ran a simple grid search for each algorithm to find a well performing setting for each (see details for grid searches below). We used the same hyperparameters across all of the tasks (i.e. so that nothing is tuned per-task). Thus, it should be possible to improve performance on a {\em given} task by tuning parameters with respect to performance for that specific task.  For these reasons, the results are not presented as upper bounds for performance with these algorithms, but rather as a starting point for comparison. It is also worth noting that we have not made a concerted effort to maximise data efficiency, for example by making many mini-batch updates using the replay buffer per step in the environment, as in \citet{popov2017data}.

The following pseudocode block demonstrates how to load a single task in the benchmark suite, run a single episode with a random agent, and compute the reward as we do for the results reported here.  Note that we run each environment for 1000 time steps and sum the rewards provided by the environment after each call to \mono{step}.  Thus, the maximum possible score for any task is 1000.  For many tasks, the practical maximum is significantly less than 1000 since it may take many steps until it's possible to drive the system into a state that gives a full reward of 1.0 each time step.

\begin{lstlisting}[language=Python]
from dm_control import suite

env = suite.load(domain_name, task_name)

spec = env.action_spec()
time_step = env.reset()
total_reward = 0.0
for _ in range(1000):
  action = np.random.uniform(spec.minimum, spec.maximum, spec.shape)
  time_step = env.step(action)
  total_reward += time_step.reward
\end{lstlisting}

\begin{figure}[p]
    \thisfloatpagestyle{empty}
    \vspace{-1cm}
    \centering
    \includegraphics[width=\textwidth]{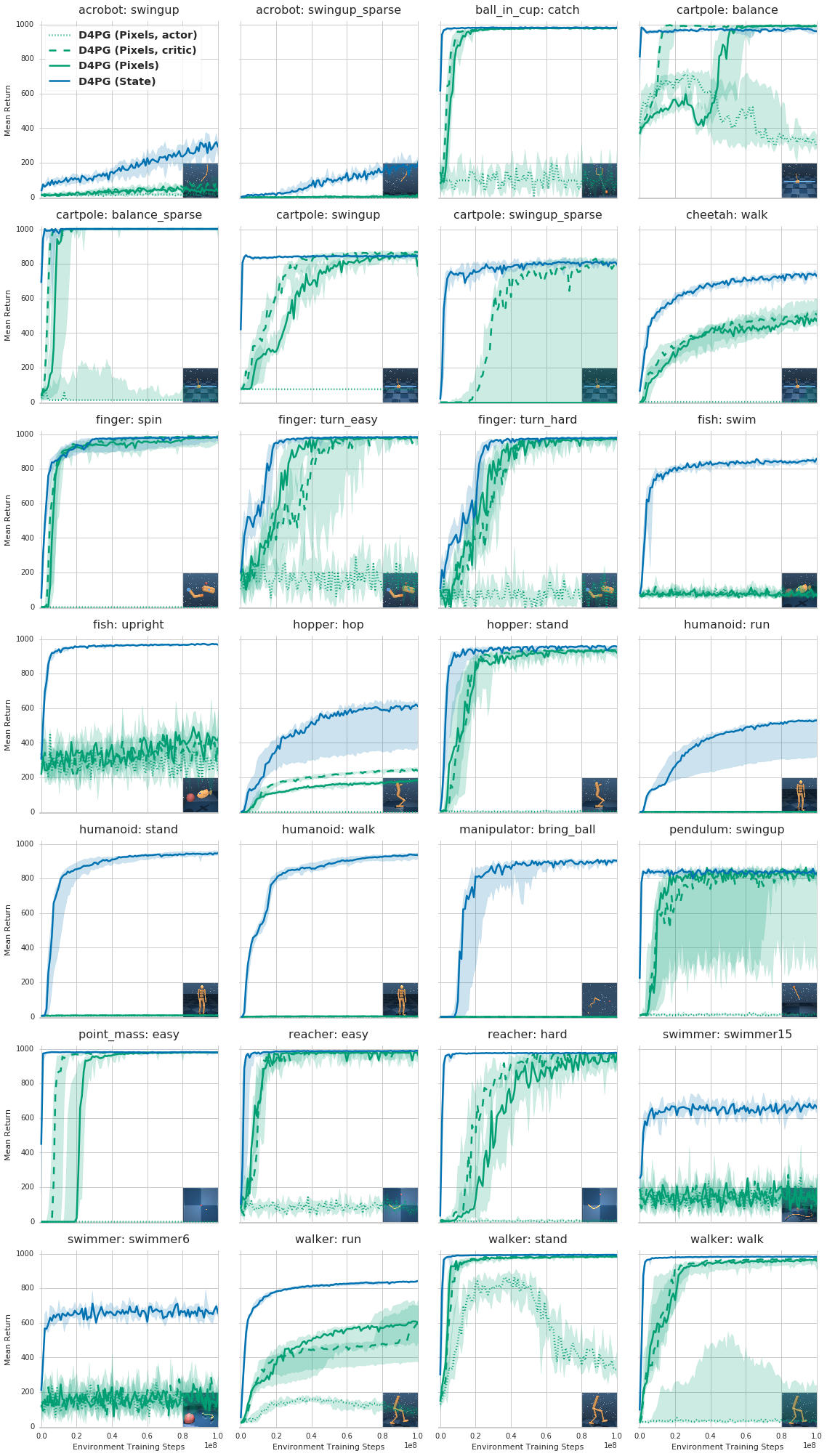}
    \caption{D4PG agent variants using pixel-only features over environment steps.}
    \label{fig:pixels-step}
\end{figure}

In the state feature case we ran 15 different seeds for each task with A3C and DDPG; for results with D4PG, which was generally found to be more stable, we ran 5 seeds.  In the raw-pixel case we also ran 5 different seeds for each task. The seed sets the network weight initialisation for the associated run.  In all cases, initial network weights were sampled using standard TensorFlow initialisers.  In the figures showing performance on individual tasks (Figures~\ref{fig:curves-step}-\ref{fig:pixels-time}), the lines denote the median performance and the shaded regions denote the 5\textsuperscript{th} and 95\textsuperscript{th} percentiles across seeds.  In the tables showing performance for individual tasks (Tables~\ref{tab:step}~\&~\ref{tab:time}) we report means and standard errors across seeds.

\begin{figure}[ht]
    \centering
    \includegraphics[width=0.9\textwidth]{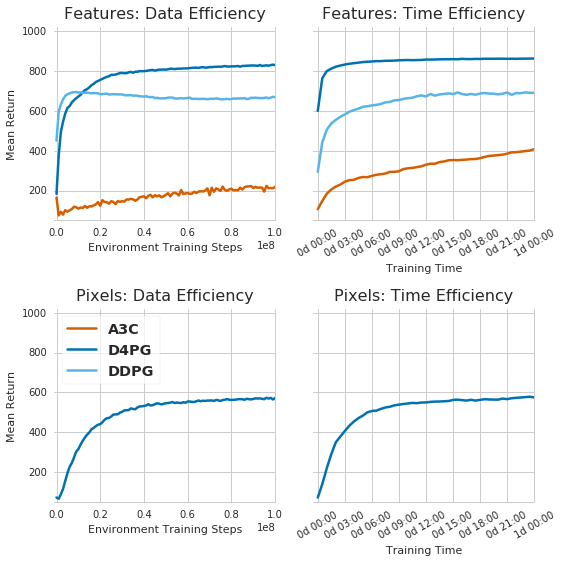}
    \caption{Mean return across all tasks in the Control Suite plotted versus data (first column) and wallclock time (second column).  The first row shows performance for A3C, DDPG and D4PG on the tasks using low-dimensional features as input.  The second row shows the performance for D4PG on the tasks using only raw-pixels as input.}
    \label{fig:aggregated}
\end{figure}

As well as studying performance on individual tasks, we examined the performance of algorithms across all tasks by plotting a simple aggregate measure.  Figure~\ref{fig:aggregated} shows the mean performance over environment steps and wallclock time for both state features and raw-pixels.  These measures are of particular interest: they offer a view into the generality of a reinforcement learning algorithm. In this aggregate view, it is clear that D4PG is the best performing agent in all metrics, with the exception that DDPG is more data efficient before $1\mathrm{e}{7}$ environment steps.  It is worth noting that the data efficiency for D4PG can be improved over DDPG by simply reducing the number of actor threads for D4PG (experiments not shown here), since with 32 actors D4PG is somewhat wasteful of environment data (with the benefit of being more efficient in terms of wall-clock).  

While we have made a concerted effort to ensure reproducible benchmarks, it's worth noting that there remain uncontrolled aspects that introduce variance into the evaluations.  For example, some tasks have a randomly placed target or initialisation of the model, and the sequence of these are not fixed across runs.  Thus, each learning run will see a different sequence of episodes, which will lead to variance in performance.  This might be fixed by introducing a fixed sequence of initialisation for episodes, but this is not in any case a practical solution for the common case of parallelised training, so our benchmarks simply reflect variability in episode initialisation sequence.

\subsection*{Algorithm and Architecture Details}

\begin{description}[leftmargin=.3cm,itemindent=-.3cm]
\item[A3C]
\cite{mnih2016asynchronous} proposed a version of the Advantage Actor Critic (A2C; \citealt{williams1991function}) that could be trained asynchronously (A3C). Here we report results for the A3C trained with 32 workers per task. The network consisted of 2 MLP layers shared between actor and critic with 256 units in the first hidden layer. The grid search explored: learning rates,  $\eta \in$ [1e{-2}, 1e{-3}, 1e{-4}, 1e{-5}, 3e{-5}, 4e{-5}, 5e{-5}]; unroll length $t_{\mathrm{max}}\in[20, 100, 1000]$; activation functions for computing \allowbreak{$\log\sigma(\cdot) \in [\mathrm{Softplus}(x), \exp(\log(0.01 + 2\text{sigmoid}(x)))]$}; number of units in the second hidden layer $\in [128,256]$; annealing of learning rate $\in [\mathrm{true}, \mathrm{false}]$.   The advantage baseline was computed using a linear layer after the second hidden layer. Actions were sampled from a multivariate Gaussian with diagonal covariance, parameterized by the output vectors $\mathbf{\mu}$ and $\mathbf{\sigma}^2$. The value of the logarithm of $\mathbf{\sigma}$ was computed using the second sigmoid activation function given above (which was found to be more stable than the $\mathrm{Softplus}(x)$ function used in the original A3C manuscript), while $\mathbf{\mu}$ was computed from a hyperbolic tangent, both stemming from the second MLP layer. The RMSProp optimiser was used with a decay factor of $\alpha=0.99$, a damping factor of $\epsilon=0.1$ and a learning rate starting at $5\mathrm{e}{-5}$ and annealed to $0$ throughout training using a linear schedule, with no gradient clipping. An entropy regularisation cost weighted at $\beta=3\mathrm{e}{-3}$ was added to the policy loss.

\item[DDPG]
\cite{lillicrap2015continuous} presented a Deep Deterministic Policy Gradients (DDPG) agent that performed well on an early version of the Control Suite. Here we present performance for straightforward single actor/learner implementation of the DDPG algorithm.
Both actor and critic networks were MLPs with ReLU nonlinearities.
The actor network had two layers of $300 \rightarrow 200$ units respectively, while the critic network had two layers of $400 \rightarrow 300$ units. The action vector was passed through a linear layer and summed up with the activations of the second critic layer in order to compute Q values.
The grid search explored:
discount factors, $\lambda \in [0.95,0.99]$;
learning rates, $\eta \in [1\mathrm{e}{-2}, 1\mathrm{e}{-3},  1\mathrm{e}{-4}, 1\mathrm{e}{-5}]$ fixed to be the same for both networks;
damping and spread parameters for the Ornstein–Uhlenbeck process, $\theta \in [0,0.15,0.85,1]$ and $\mu \in [0.1, 0.2, 0.3, 0.4]$ respectively; hard (swap at intervals of 100 steps) versus soft ($\tau=1\mathrm{e}{-3}$)  target updates.
For the results shown here the two networks were trained with independent Adam optimisers~\cite{kingma2014adam}, both with a learning rate of $\eta=1\mathrm{e}{-4}$, with gradients clipped at $[-1, 1]$ for the actor network. The agent used discounting of $\lambda=0.99$. As in the paper, we used a target network with soft updates and an Ohrstein-Uhlenbeck process to add an exploration noise that is correlated in time, with similar parameters, except for a slightly bigger $\sigma$ ($\theta= 0.15$, $\sigma = 0.3$, $\tau = 1\mathrm{e}{-3}$). The replay buffer size was also kept to $1\mathrm{e}6$, and training was done with a minibatch size of $64$.

\item[D4PG]
The Distributional Distributed Deep Deterministic Policy Gradients algorithm~\citep{d4pg} extends regular DDPG with the following features: First, the critic value function is modelled as a categorical distribution \citep{bellemare2017distributional}, using 101 categories spaced evenly across $[-150,150]$. Second, acting and learning are decoupled and deployed on separate machines using the Ape-X architecture described in \citep{apex}. We used 32 CPU-based actors and a single GPU-based learner for benchmarking. D4PG additionally applies $N$-step returns with $N=5$, and non-uniform replay sampling \citep{schaul2015prioritized} ($\alpha_{\text{sample}}=0.6$) and eviction ($\alpha_{\text{evict}}=0.6$) strategies using a sample-based distributional KL loss (see \citep{apex} and \citep{d4pg} for details). D4PG hyperparameters were the same as those used for DDPG, with the exception that (1) hard target network updates are applied every 100 steps, and (2) exploration noise is sampled from a Gaussian distribution with fixed $\sigma$ varying from $1/32$ to $1$ across the actors. A mini-batch size of $256$ was used.
\end{description}

\subsection*{Results: Learning from state features}

\begin{figure}[p]
    \thisfloatpagestyle{empty}
    \vspace{-1cm}
    \centering
    \caption{Comparison of A3C, DDPG, D4PG agents over environment steps.}
    \includegraphics[width=\textwidth]{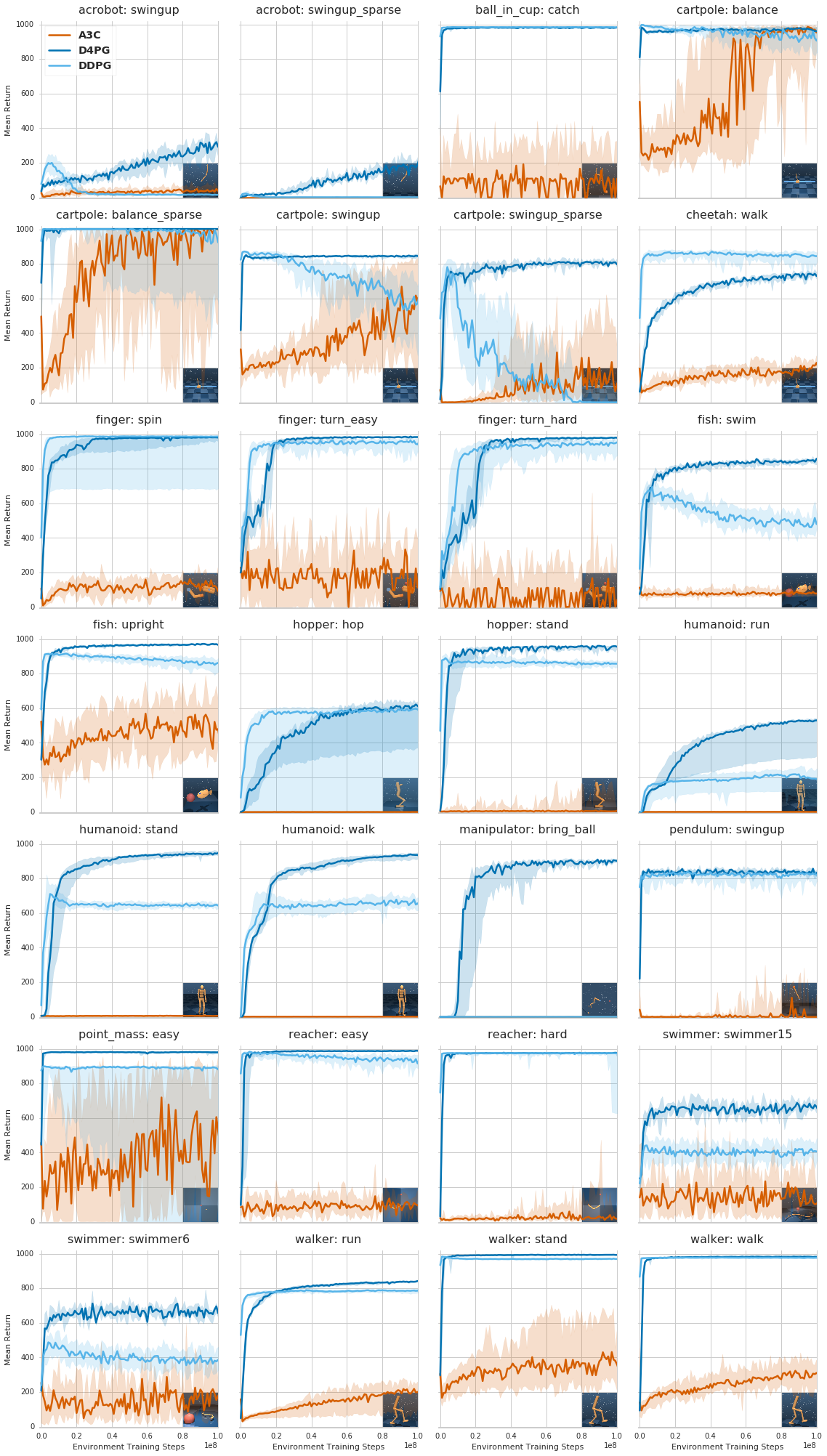}
    \label{fig:curves-step}
\end{figure}

Due to the different parallelization architectures, the evaluation protocol for each agent was slightly different: DDPG was evaluated for 10 episodes for every 100000 steps (with no exploration noise), and A3C was trained with 32 workers and concurrently evaluated with another worker that updated its parameters every 3 episodes, which produced intervals of on average 96000 steps per update. The plots in Figure~\ref{fig:curves-step} and Figure~\ref{fig:curves-time} show the median and the 5th and 95th percentile of the returns for the first $1\mathrm{e}8$ steps. Each agent was run 15 times per task using different seeds (except for D4PG which was run 5 times), using only low-dimensional state feature information. D4PG tends to achieve better results in nearly all of the tasks. Notably, it manages to reliably solve the \mono{manipulator:bring\_ball} task, and achieves a good performance in \mono{acrobot} tasks. We found that part of the reason the agent did not go above $600$ in the \mono{acrobot} task is due to the time it takes for the pendulum to be swung up, so its performance is probably close to the upper bound.

\subsection*{Results: Learning from pixels}

The \textit{DeepMind Control Suite} can be configured to produce observations containing any combination of state vectors and pixels generated from the provided cameras. We also benchmarked a variant of D4PG that learns directly from pixel-only input, using $84\times84$ RGB frames from the $0^\mathrm{th}$ camera.  To process pixel input, D4PG is augmented with a simple 2-layer ConvNet. Both kernels are size $3\times3$ with 32 channels and ELU activation, and the first layer has stride 2. The output is fed through a single fully-connected layer with 50 neurons, with layer normalisation~\cite{ba2016layer} and tanh() activations. We explored four variants of the algorithm.  In the first, there were separate networks for the actor and Q-critic.  In the other three, the actor and critic shared the convolutional layers, and the actor and critic each had a separate fully connected layer before their respective outputs. The best performance was obtained by weight-sharing the convolutional kernel weights between the actor and critic networks, and only allowing these weights to be updated by the critic optimiser (i.e. truncating the policy gradients after the actor MLP). D4Pixels internally frame-stacks 3 consecutive observations as the ConvNet input.

\begin{figure}[p]
    \thisfloatpagestyle{empty}
    \vspace{-1cm}
    \centering
    \includegraphics[width=\textwidth]{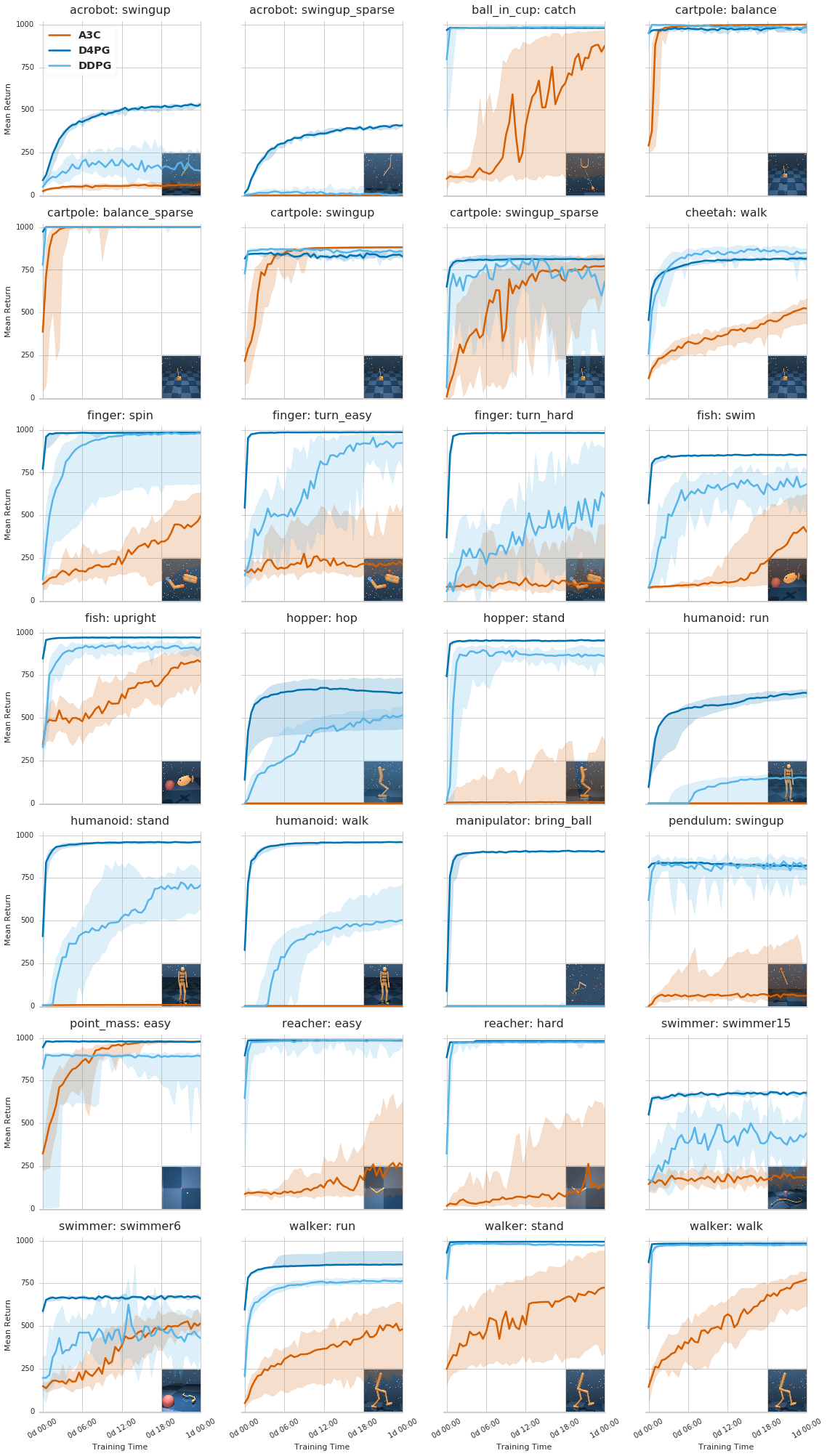}
    \caption{Comparison of A3C, DDPG and D4PG agents over 1 day of training time.}
    \label{fig:curves-time}
\end{figure}

\begin{figure}[p]
    \thisfloatpagestyle{empty}
    \vspace{-1cm}
    \centering
    
    \includegraphics[width=\textwidth]{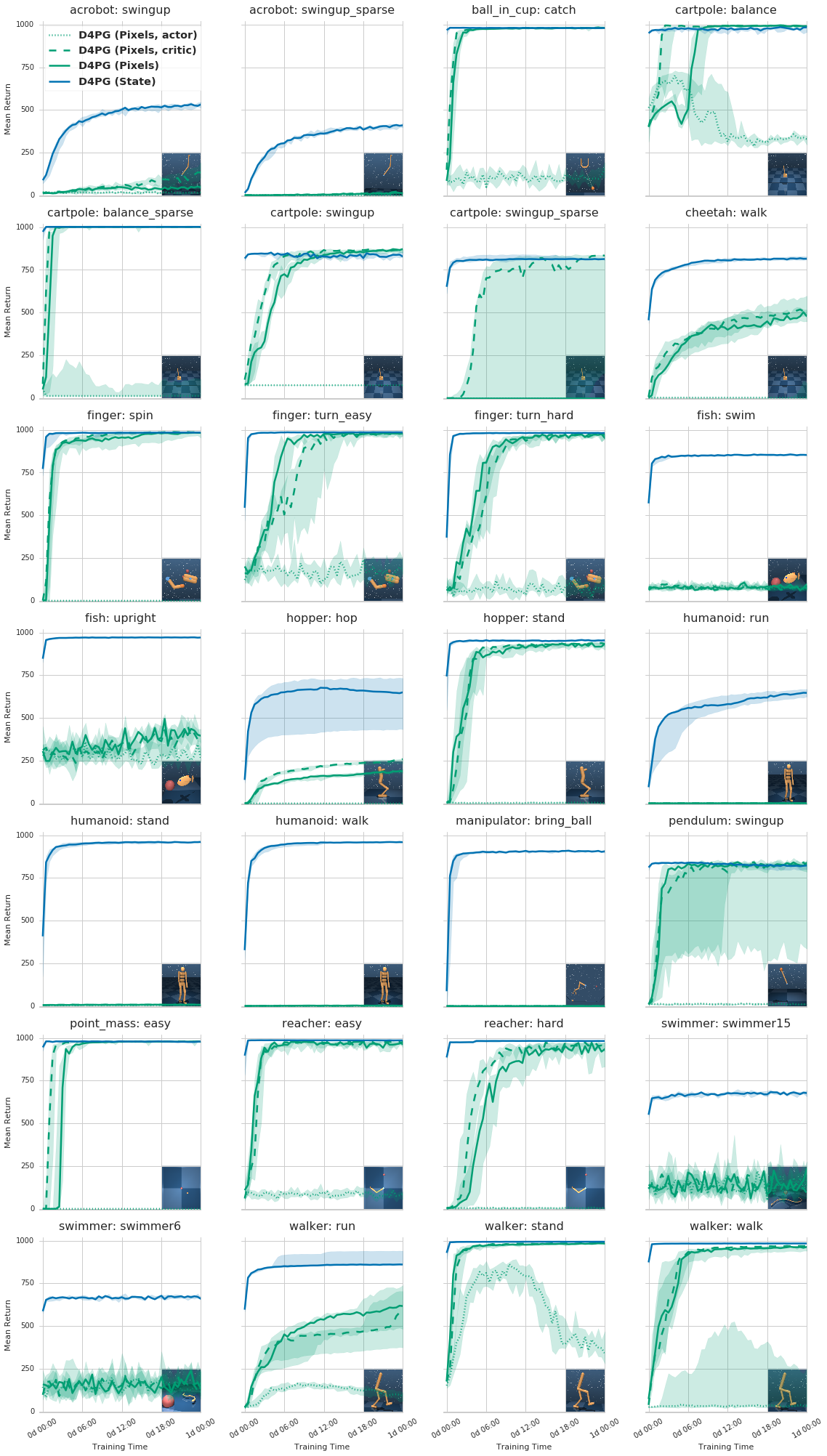}
    \caption{D4PG agent variants using pixel-only features over 1 day of training time.}
    \label{fig:pixels-time}
\end{figure}

Results for 1 day of running time are shown in Figure~\ref{fig:pixels-time}; we plot the results for the three shared-weights variants of D4PG, with gradients into the ConvNet from the actor (dotted green), critic (dashed green), or both (solid green). For the sake of comparison, we plot D4PG performance for low-dimensional features (solid blue) from Figure~\ref{fig:curves-time}. The variant that employed separate networks for actor and critic performed significantly worse than the best of these and is not shown. Learning from pixel-only input is successful on many of the tasks, but fails completely in some cases. It is worth noting that the camera view for some of the task domains are not well suited to a pixel-only solution for the task. Thus, some of the failure cases are likely due to the difficulty of positioning a camera that simultaneously captures both the navigation targets as well as the details of the agents body: e.g., in the case of swimmer:swimmer6 and swimmer15 as well as fish:swim.

\section{Conclusion and future work}

The \textit{DeepMind Control Suite} is a starting place for the design and  performance comparison of reinforcement learning algorithms for physics-based control. It offers a wide range of tasks, from near-trivial to quite difficult.  The uniform reward structure allows for robust suite-wide performance measures.

The results presented here for A3C, DDPG, and D4PG constitute baselines using, to the best of our knowledge, well performing implementations of these algorithms.  At the same time, we emphasise that the learning curves are not based on exhaustive hyperparameter optimisation, and that for a given algorithm the same hyperparameters were used across all tasks in the Control Suite.  Thus, we expect that it may be possible to obtain better performance or data efficiency, especially on a per-task basis.

We are excited to be sharing the Control Suite with the wider community and hope that it will be found useful. We look forward to the diverse research the Suite may enable, and to integrating community contributions in future releases.

\subsection*{Future work}
Several elements are missing from the current release of the Control Suite. 

Some features, like the lack of rich tasks, are missing by design. The Suite, and particularly the \mono{benchmarking} set of tasks, is meant to be a stable, simple starting point for learning control. Task categories like full manipulation and locomotion in complex terrains require reasoning about a distribution of tasks and models, not only initial states. These require more powerful tools which we hope to share in the future in a different branch.

There are several features that we hoped to include but did not make it into this release; we intend to add these in the future. They include: a quadrupedal locomotion task, an interactive visualiser with which to view and perturb the simulation, support for C callbacks and multi-threaded dynamics, a MuJoCo TensorFlow op wrapper and Windows{\scriptsize ™} support.

\begin{table}[p]
    \small
    \centerline{\begin{tabular}{rrrrrr}
\toprule
        &         &          A3C &          D4PG & D4PG (Pixels) &          DDPG \\
Domain & Task &              &               &                       &               \\
\midrule
acrobot & swingup &   41.9 ± 1.2 &   297.6 ± 8.4 &            81.7 ± 4.4 &    15.4 ± 0.9 \\
        & swingup\_sparse &    0.2 ±  0.1 &   198.7 ± 9.1 &            13.0 ± 1.7 &     1.0 ± 0.1 \\
ball\_in\_cup & catch &  104.7 ± 7.8 &   981.2 ± 0.7 &           980.5 ± 0.5 &   984.5 ± 0.3 \\
cartpole & balance &  951.6 ± 2.4 &   966.9 ± 1.9 &           992.8 ± 0.3 &   917.4 ± 2.2 \\
        & balance\_sparse &  857.4 ± 7.9 &  1000.0 ± 0.0 &          1000.0 ± 0.0 &   878.5 ± 8.0 \\
        & swingup &  558.4 ± 6.8 &   845.5 ± 1.2 &           862.0 ± 1.1 &   521.7 ± 6.1 \\
        & swingup\_sparse &  179.8 ± 5.9 &   808.4 ± 4.4 &          482.0 ± 56.6 &     4.5 ± 1.1 \\
cheetah & walk &  213.9 ± 1.6 &   736.7 ± 4.4 &           523.8 ± 6.8 &   842.5 ± 1.6 \\
finger & spin &  129.4 ± 1.5 &   978.2 ± 1.5 &           985.7 ± 0.6 &   920.3 ± 6.3 \\
        & turn\_easy &  167.3 ± 9.6 &   983.4 ± 0.6 &           971.4 ± 3.5 &   942.9 ± 4.3 \\
        & turn\_hard &   88.7 ± 7.3 &   974.4 ± 2.8 &           966.0 ± 3.4 &   939.4 ± 4.1 \\
fish & swim &   81.3 ± 1.1 &   844.3 ± 3.1 &            72.2 ± 3.0 &   492.7 ± 9.8 \\
        & upright &  474.6 ± 6.6 &   967.4 ± 1.0 &          405.7 ± 19.6 &   854.8 ± 3.3 \\
hopper & hop &    0.5 ± 0.0 &  560.4 ± 18.2 &           242.0 ± 2.1 &   501.4 ± 6.0 \\
        & stand &   27.9 ± 2.3 &   954.4 ± 2.7 &           929.9 ± 3.8 &   857.7 ± 2.2 \\
humanoid & run &    1.0 ± 0.0 &  463.6 ± 13.8 &             1.4 ± 0.0 &   167.9 ± 4.1 \\
        & stand &    6.0 ± 0.1 &   946.5 ± 1.8 &             8.6 ± 0.2 &   642.6 ± 2.1 \\
        & walk &    1.6 ± 0.0 &   931.4 ± 2.3 &             2.6 ± 0.1 &   654.2 ± 3.9 \\
manipulator & bring\_ball &    0.4 ± 0.0 &   895.9 ± 3.7 &             0.5 ± 0.1 &     0.6 ± 0.1 \\
pendulum & swingup &   48.6 ± 5.2 &   836.2 ± 5.0 &          680.9 ± 41.9 &   816.2 ± 4.7 \\
point\_mass & easy &  545.3 ± 9.3 &   977.3 ± 0.6 &           977.8 ± 0.5 &  618.0 ± 11.4 \\
reacher & easy     &   95.6 ± 3.5 &   987.1 ± 0.3 &           967.4 ± 4.1 &   917.9 ± 6.2 \\
        & hard &   39.7 ± 2.9 &   973.0 ± 2.0 &           957.1 ± 5.4 &   904.3 ± 6.8 \\
swimmer & swimmer15 &  164.0 ± 7.3 &  658.4 ± 10.0 &          180.8 ± 11.9 &  421.8 ± 13.5 \\
        & swimmer6 &  177.8 ± 7.8 &  664.7 ± 11.1 &          194.7 ± 15.9 &  394.0 ± 14.1 \\
walker & run &  191.8 ± 1.9 &   839.7 ± 0.7 &          567.2 ± 18.9 &   786.2 ± 0.4 \\
        & stand &  378.4 ± 3.5 &   993.1 ± 0.3 &           985.2 ± 0.4 &   969.8 ± 0.3 \\
        & walk &  311.0 ± 2.3 &   982.7 ± 0.3 &           968.3 ± 1.8 &   976.3 ± 0.3 \\
\bottomrule
\end{tabular}}
    \caption{Mean and Standard Error of 100 episodes after $10^8$ training steps for each seed.}
    \label{tab:step}
\end{table}

\begin{table}[p]
    \thisfloatpagestyle{empty}
    \small
    \centerline{\begin{tabular}{rrrrrr}
\toprule
        &         &           A3C &          D4PG & D4PG (Pixels) &          DDPG \\
Domain & Task &               &               &                       &               \\
\midrule
acrobot & swingup &    64.7 ± 2.0 &   531.1 ± 8.9 &           141.5 ± 6.9 &   149.8 ± 4.2 \\
        & swingup\_sparse &     0.3 ± 0.1 &   415.9 ± 9.5 &            16.3 ± 2.0 &     8.5 ± 0.8 \\
ball\_in\_cup & catch &  643.3 ± 10.9 &   979.8 ± 0.7 &           980.5 ± 0.6 &   984.4 ± 0.3 \\
cartpole & balance &   999.0 ± 0.0 &   982.7 ± 1.4 &           988.3 ± 0.6 &   980.2 ± 0.6 \\
        & balance\_sparse &  999.7 ± 0.1 &  999.7 ± 0.2 &          1000.0 ± 0.0 &   998.4 ± 0.7 \\
        & swingup &   881.4 ± 0.1 &   836.1 ± 1.6 &           864.0 ± 1.0 &   855.6 ± 0.6 \\
        & swingup\_sparse &   752.5 ± 3.6 &   814.6 ± 0.7 &          649.6 ± 46.9 &   591.1 ± 7.3 \\
cheetah & walk &   519.7 ± 2.2 &   814.6 ± 4.0 &           524.7 ± 6.8 &   849.7 ± 1.7 \\
finger & spin &   446.7 ± 4.2 &   984.7 ± 0.4 &           986.2 ± 0.4 &   916.1 ± 6.3 \\
        & turn\_easy &  291.9 ± 11.3 &   986.1 ± 0.5 &           976.0 ± 1.7 &   927.8 ± 4.9 \\
        & turn\_hard &   200.6 ± 9.6 &   983.2 ± 0.5 &           971.0 ± 2.9 &  618.7 ± 11.8 \\
fish & swim &   395.7 ± 7.3 &   852.5 ± 2.7 &            76.8 ± 4.2 &   666.8 ± 6.8 \\
        & upright &   813.6 ± 4.4 &   971.1 ± 0.9 &          366.8 ± 18.3 &   915.3 ± 2.0 \\
hopper & hop &     0.7 ± 0.1 &  613.2 ± 18.0 &           249.8 ± 2.1 &   435.2 ± 4.9 \\
        & stand &    82.6 ± 5.9 &   956.8 ± 3.9 &           930.0 ± 3.4 &   862.8 ± 2.8 \\
humanoid & run &     1.0 ± 0.0 &   643.1 ± 3.1 &             1.5 ± 0.0 &   137.3 ± 1.3 \\
        & stand &     7.8 ± 0.1 &   959.1 ± 0.8 &             8.7 ± 0.2 &   687.9 ± 2.1 \\
        & walk &     1.7 ± 0.0 &   960.3 ± 0.5 &             2.7 ± 0.1 &   534.4 ± 2.1 \\
manipulator & bring\_ball &     0.6 ± 0.1 &   903.7 ± 3.9 &             0.6 ± 0.1 &     1.2 ± 0.4 \\
pendulum & swingup &    98.8 ± 9.5 &   814.9 ± 5.8 &          705.7 ± 36.7 &   818.6 ± 4.3 \\
point\_mass & easy &   978.5 ± 0.1 &   978.2 ± 0.7 &           979.3 ± 0.3 &   862.3 ± 3.1 \\
reacher & easy     &   285.3 ± 8.9 &   984.9 ± 1.1 &           967.1 ± 4.4 &   969.8 ± 3.0 \\
        & hard &   247.5 ± 9.4 &   982.8 ± 0.4 &           958.0 ± 5.2 &   975.3 ± 0.5 \\
swimmer & swimmer15 &   196.8 ± 8.4 &   681.1 ± 9.3 &          146.0 ± 12.0 &  410.9 ± 10.5 \\
        & swimmer6 &   526.4 ± 9.6 &  651.0 ± 10.0 &          168.9 ± 13.1 &  461.0 ± 10.6 \\
walker & run &   448.4 ± 4.1 &   891.5 ± 5.6 &          566.1 ± 19.1 &   766.5 ± 0.4 \\
        & stand &   707.9 ± 6.5 &   994.0 ± 0.3 &           984.8 ± 0.4 &   973.7 ± 0.3 \\
        & walk &   744.2 ± 2.4 &   983.1 ± 0.4 &           968.1 ± 1.8 &   975.7 ± 0.3 \\
\bottomrule
\end{tabular}
}
    \caption{Mean and standard error of 100 episodes after 24 hours of training for each seed.}
     \label{tab:time}
\end{table}

\newpage

{\small
\bibliographystyle{plainnat}
\bibliography{biblio}
}

\end{document}